\documentclass[letterpaper,10pt,conference]{ieeeconf}

\IEEEoverridecommandlockouts
\usepackage[T1]{fontenc}
\usepackage{graphicx}
\usepackage{makecell}
\usepackage{subcaption}
\usepackage{float}
\usepackage{placeins}
\usepackage{adjustbox}
\usepackage{amsmath}
\usepackage{amssymb}
\usepackage{booktabs}
\usepackage{graphicx}
\usepackage{tikz}
\usetikzlibrary{arrows.meta,positioning,fit,shadings,shapes.geometric}
\usepackage{xcolor}
\usepackage{xspace}
\usepackage{microtype}
\usepackage{hyperref}
\usepackage{cleveref}
\hypersetup{hidelinks}

\newcommand{\method}{\textsc{Metatron}\xspace}

\newif\ifincludeappendix
\includeappendixtrue

\title{\LARGE \bf
\method: A Promptable Whole-Body Execution Layer for Humanoids
}

\author{Chen Tessler$^{1,*}$, Yifeng Jiang$^{1,*}$, Alec Otto$^{2}$, Yi Shi$^{1,\dagger}$,\\
Umar Iqbal$^{1}$, Gal Chechik$^{1}$, Xue Bin Peng$^{1}$%
\thanks{$^{1}$NVIDIA. $^{2}$Dollhouse Robotics. $^{*}$Equal contribution.
$^{\dagger}$Work done while interning at NVIDIA.}}

\IEEEaftertitletext{% Naming metaphor, not a verbatim quotation. Used inside the title-area hook.
\noindent\makebox[\textwidth][c]{%
\begin{minipage}{0.85\textwidth}
\centering\small\itshape
Metatron bridges divine will and the physical world---\\
an intermediary through whom intent becomes action.
\par\smallskip
{\normalfont\footnotesize Inspired by Orlov (2005) and Scholem (1941)~\cite{orlov2005enoch,scholem1941major}.}
\par\smallskip
{\normalfont\footnotesize Interactive demo: \url{https://nvlabs.github.io/ProtoMotions/demos/metatron/}}
\par\medskip
\end{minipage}%
}
}

\begin{document}

\maketitle
\thispagestyle{empty}
\pagestyle{empty}

\begin{abstract}
% Humanoid controllers often require task-specific training despite sharing the need for balance and whole-body coordination. We present \method, a promptable whole-body execution layer combining a generative base policy with a programmable skill layer. The base policy accepts partial prompts, such as timed body-part targets, velocity commands, style cues, or any subset, and produces diverse whole-body motions consistent with the specified goals. Training combines motion capture with synthetic recovery trajectories, and retargeted scene-interaction motions. After training, the policy is frozen. Stateful programs called skills convert task intent into prompts at each control step, drawing rough geometric sketches that the policy renders into coordinated movement. Skills are ordinary code, refined by coding agents that analyze motion data and fit compact rules without any task-specific training of the policy. On the Unitree G1, one frozen checkpoint supports walking, recovery, teleoperated object pickup, and sitting on furniture, outperforming reference-tracking pipelines. The same recipe produces climbing and acrobatics on a simulated character. Switching tasks requires only changing the prompt.

Humanoid controllers often require task-specific training despite sharing the need for balance and whole-body coordination. We present \method, a promptable whole-body execution layer combining a generative base policy with a programmable skill layer. The base policy receives partial goals, collectively called a prompt, that can combine timed body-part targets, velocity commands, and style cues. It produces diverse whole-body motions consistent with these goals. Training combines motion capture, synthetic recovery trajectories, and retargeted scene-interaction motions. After training, the policy is frozen. Stateful programs called skills express task intent as prompts at each control step, drawing rough geometric sketches that the policy turns into coordinated movement. Skills are ordinary code, refined by coding agents that analyze motion data and fit compact rules without task-specific policy training. On the physical Unitree G1, one frozen checkpoint supports walking, recovery, teleoperated object pickup, and sitting on furniture. Comparative evaluations show higher reaching success and push robustness than the tested reference-tracking pipelines. The same recipe produces climbing and acrobatics on a simulated character. \method enables coding agents to turn learned motor capabilities into new skills through code, without further policy training.
\end{abstract}

\section{Introduction}
\label{sec:introduction}

Consider a humanoid that must walk to a table in a particular style, grasp a cup at a designated moment, recover after a shove, present it to a human, and shake hands. Today this typically requires separate learned controllers for locomotion~\cite{gu2024humanoid}, manipulation~\cite{fu2024humanplus}, recovery~\cite{he2025humanup}, compliant control~\cite{margolis2025softmimic}, and motion tracking~\cite{liao2026beyondmimic}. These controllers duplicate capabilities such as balance and coordination, and transitions between them demand additional engineering. 

We argue that a more scalable recipe is to separate \emph{what} a motion should accomplish from \emph{how} to coordinate the body: train one general-purpose whole-body policy with a flexible goal-conditioning interface, and let task-specific programs specify the goals. The policy becomes an execution layer that provides physical coordination; the programs become the controllers that supply intent. Each program specifies a partial motion goal that we call a \emph{prompt}: a sitting command specifies the shape and location of a nearby object, while a style cue biases the generated motion toward example poses.

\begin{figure}[!t]
    \centering
    \includegraphics[trim={1.5cm 0cm 1.5cm 0cm},clip,width=0.99\linewidth]{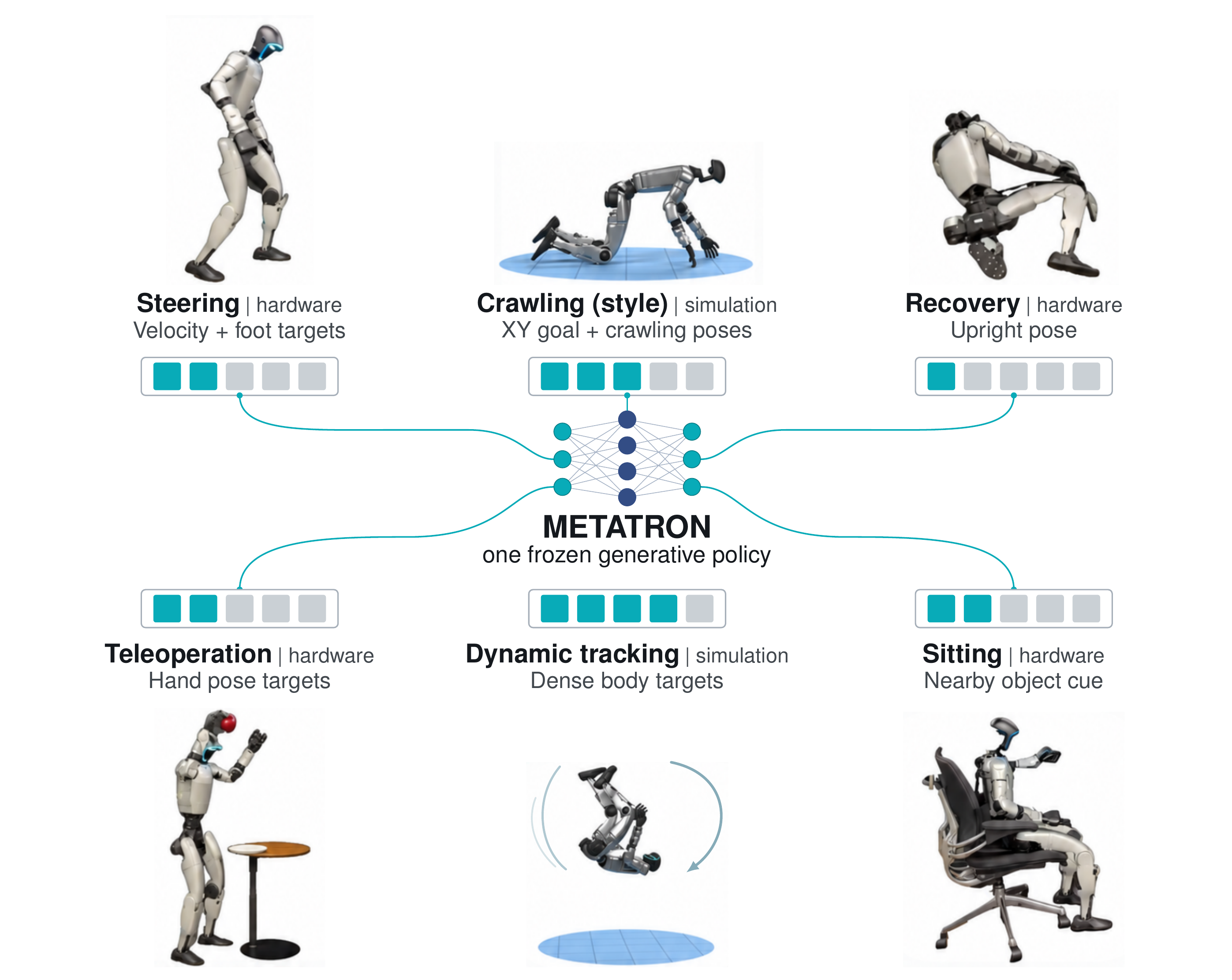}
    % \caption{\textbf{One frozen controller, diverse behaviors.} Partial goals specify selected aspects of the desired motion, while the policy provides whole-body coordination. Changing the prompts enables steering, recovery, teleoperated object pickup, and sitting on G1, as well as dynamic motion tracking and style-conditioned crawling in simulation. Teal blocks indicate specified goal components; gray blocks indicate those left unspecified. The policy weights remain fixed across behaviors, with no task-specific policy training.}
    \caption{\textbf{One frozen controller per embodiment, diverse behaviors.} Partial prompts specify goals; each controller provides whole-body coordination across tasks without task-specific training. Teal and gray blocks indicate specified and unspecified goal components, respectively.}
    \label{fig: teaser}
\vspace{-0.6cm}
\end{figure}

We present \method, a promptable whole-body execution layer for humanoids (\Cref{fig: teaser}). It combines a capable base policy with a programmable skill layer. The base policy builds on MaskedMimic~\cite{tessler2024maskedmimic}, which trains a dense motion tracker and distills it into a generative policy that accepts sparse body-part targets at specified future times. 

Once trained, the policy is frozen and a runtime layer runs programs called \emph{skills} that convert task intent into prompts at each control step (\Cref{sec:prompt-compiler}). Skills are ordinary code, authored by humans or coding agents. We show how LLM agents analyze motion clips and fit compact rules that capture shared motion characteristics, turning task design into a software problem. Reinforcement learning trains the motion experts; on-policy distillation teaches the base policy to imitate them under partial goals, including at states reached by the student itself. This training supports approximate goal sketches rather than requiring exact reference trajectories.
%Importantly, because the base policy is trained with reinforcement learning across a large and randomized state distribution, the prompt need not be exact: the policy generalizes to approximate goals and recovers from errors that imperfect commands introduce. This tolerance for imprecision is what allows task programs to be compact rules rather than additional learned models.

\method also extends MaskedMimic in several directions. Forward kinematics reconstructs local body poses from joint and inertial measurements, without external motion capture; world-position goals additionally require localization. Scene elements are represented as geometric primitives specified by measured dimensions and poses. Style conditioning allows operators and skills to select the generated motion's character. We augment the training data with synthetic push-recovery and get-up trajectories, extending the repertoire beyond what motion capture alone provides. On the modeling side, spherical projection bounds the stochastic latent, while predictive and temporal auxiliary losses regularize its fused representation (\Cref{sec:distillation}).

Our contributions are a whole-body policy with a flexible prompt interface, training data augmented with scene interactions and synthetic recoveries (\Cref{sec:training-data}), and a skill layer whose code can be refined by coding agents without policy retraining (\Cref{sec:prompt-compiler}). One frozen checkpoint per embodiment supports the demonstrated tasks on physical G1 and simulated SOMA. Separate G1 comparisons show higher whole-body reaching success and push robustness than the tested reference-tracking pipelines (\Cref{sec:experiments}).

\section{Related Work}
\label{sec:related}

\textbf{Physics-based motion imitation:}
Reference tracking has progressed from motion-specific policies
\cite{peng2018deepmimic} to controllers that follow large and diverse motion
collections \cite{luo2023phc,chen2025gmt,luo2025sonic}. Dense tracking provides
a strong training signal for coordinated motion, but a tracking interface still
requires a reference trajectory. \method uses this supervision to distill a policy that accepts partial goals.

\textbf{Partial-goal and generalist motion control:}
Latent generative controllers such as ASE, CALM, and PULSE learn reusable
motion representations \cite{peng2022ase,tessler2023calm,luo2024pulse}.
GPC \cite{shi2026gpc} learns a
generative controller over discrete motion codes. Such representations can support downstream tasks through task-specific policy training. MaskedMimic
\cite{tessler2024maskedmimic} instead
trains a controller directly on partial motion descriptions, supporting
multiple command types and transitions between them in simulation.
\method extends this approach with a larger training set including synthetic
recovery motions and retargeted scene interactions, a latent design that
separates intent from solution, a skill layer that makes task logic explicit
as ordinary code refined by coding agents, and body-pose reconstruction from
robot sensors for physical deployment
(\Cref{sec:method,sec:prompt-compiler}).
Other distillation-based controllers~\cite{he2025hover, xu2026interprior, li2026vaic} likewise
distill one policy for many command types, and concurrent generalist trackers scale a similar partial-goal
interface~\cite{yu2026omnicontact, zeng2026scaling}. Our emphasis is on combining optional goal streams with explicit skill programs: humans and coding agents refine the prompts without task-specific policy training.

\textbf{From animation to robotics:}
Large-scale motion tracking and whole-body teleoperation demonstrate that
human motion can supervise versatile robot control
\cite{chen2025gmt,luo2025sonic,he2024omnih2o}. BeyondMimic
\cite{liao2026beyondmimic} further combines tracking with latent diffusion and
test-time guidance for new objectives. Guidance adjusts generated motion to
meet an objective at runtime. Our policy instead learns to respond directly to
partial goals: skills modify
the prompt, and one prior-and-decoder pass produces joint targets at each
control step, without task-specific optimization of the generated motion.
% TODO: Add measured baseline comparisons once the evaluation is frozen.

\section{\method: Base Policy}
\label{sec:method}

\begin{figure*}[t]
\centering
\resizebox{\textwidth}{!}{%
\begin{tikzpicture}[
    font=\small,
    box/.style={draw=black!70, rounded corners=2pt, align=center, minimum height=9mm,
                minimum width=31mm, fill=white, inner sep=3pt, line width=0.5pt},
    data/.style={box, fill=gray!12},
    prompt/.style={box, fill=orange!18},
    net/.style={box, fill=blue!10},
    lat/.style={box, fill=green!12},
    train/.style={dashed, draw=red!60!black},
    arr/.style={-{Latex[length=2mm]}, line width=0.7pt, draw=black!80},
    tarr/.style={arr, dashed, draw=red!60!black},
    lbl/.style={font=\footnotesize, fill=white, inner sep=1.5pt}
]
% Columns: 4.2, 7.8, 11.4, 14.8, 18.2.
% Rows: training 4.6; prompt 3.0; history 1.5; auxiliary 0.35.

% ---------------- training-only band ----------------
\fill[red!4] (-2.7,3.95) rectangle (20.0,5.85);
\node[font=\footnotesize\itshape, text=red!60!black, anchor=north west]
    at (-2.65,5.85) {training only};

\node[data, train] (gt) at (4.2,4.6)
    {clean state $s_t$,\\full future motion $y_t$};
\node[net, train] (post) at (7.8,4.6) {posterior encoder};
\node[lat, train] (dmu) at (11.4,4.6)
    {mean correction, std.\\$\Delta\mu_t,\ \sigma_t'$};
\draw[tarr] (gt.east) -- (post.west);
\draw[tarr] (post.east) -- (dmu.west);

% ---------------- prompt row ----------------
\node[prompt] (pr) at (4.2,3.0)
    {masked prompt $g_t$\\$\mathcal C^w,\ \mathcal C^\ell,\ u,\ \xi,\ b$};
\node[net, minimum height=24mm] (enc) at (7.8,2.25)
    {prior encoder\\(transformer)};
\node[lat] (prior) at (11.4,3.0) {prior\\$(\mu_t,\sigma_t)$};
\node[circle, ball color=green!45!white, minimum size=13mm, inner sep=0]
    (z) at (14.8,3.0) {};
\draw[black!50, line width=0.4pt] (z.center) ellipse (6.5mm and 2.2mm);
\fill[black] ([shift={(1.6mm,2.9mm)}]z.center) circle (0.6mm);
\node[font=\footnotesize, align=center] at (14.8,4.0)
    {sample latent, project\\to the unit sphere};
\draw[arr] (pr.east) -- (pr.east -| enc.west);
\draw[tarr] (pr.north east) -- (post.south west);
\draw[arr] (prior.west -| enc.east) -- (prior.west);
\draw[arr] (prior.east) -- (z.west);
\draw[tarr] (gt.south) -- node[lbl, right]{mask most entries} (pr.north);
\draw[tarr] (dmu.east)
    -- node[lbl, pos=0.35, above]{$\Delta\mu_t,\ \sigma_t'$}
    (16.1,4.6) |- (z.east);
\draw[train, {Latex[length=2mm]}-{Latex[length=2mm]}, line width=0.7pt]
    (dmu.south) -- node[lbl, right]{$D_{\mathrm{KL}}(q\|p)$} (prior.north);

% ---------------- history row ----------------
\node[data] (hist) at (4.2,1.5)
    {sensor and action history\\$o_t,\ a_{t-1}$};
\node[lat] (d) at (11.4,1.5) {condition $d_t$};
\node[net] (fuse) at (14.8,1.5) {fuse $(d_t,z_t)\to h_t$};
\node[net] (dec) at (18.2,1.5) {decoder};
\draw[arr] (hist.east) -- (hist.east -| enc.west);
\draw[arr] (d.west -| enc.east) -- (d.west);
\draw[arr] (d.east) -- (fuse.west);
\draw[arr] (z.south) -- node[lbl, right]{$z_t$} (fuse.north);
\draw[arr] (fuse.east) -- (dec.west);

% ---------------- outputs and losses ----------------
\node[box] (a) at (18.2,3.0) {joint commands $a_t$};
\draw[arr] (dec.north) -- (a.south);
\node[net, train] (trk) at (18.2,4.6) {frozen tracker\\$a_t^{\mathrm{trk}}$};
\draw[tarr] (trk.south) -- node[lbl, right]{action loss} (a.north);
\node[lat, train] (aux) at (14.8,0.35)
    {auxiliary losses on $h_t^q$\\dynamics, smoothness};
\draw[tarr] (fuse.south) -- (aux.north);

% Route the history bypass just below the auxiliary-loss box.
\coordinate (bypass) at ([yshift=-2mm]aux.south);
\draw[arr, draw=black!45]
    (hist.south)
    -- (hist.south |- bypass)
    -- (dec.south |- bypass)
    -- node[lbl, midway, right, text=black!60]{$o_t,\ a_{t-1}$}
    (dec.south);

% ---------------- inputs: pose images ----------------
\node[inner sep=0] (f1) at (-1.9,4.75)
    {\includegraphics[height=14mm,width=11mm,keepaspectratio]{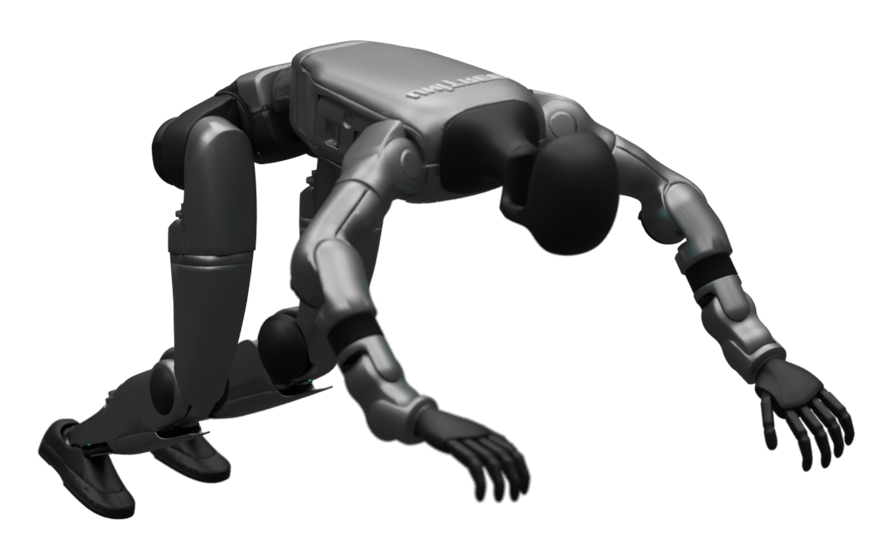}};
\node[inner sep=0] (f2) at (-0.7,4.75)
    {\includegraphics[height=14mm,width=11mm,keepaspectratio]{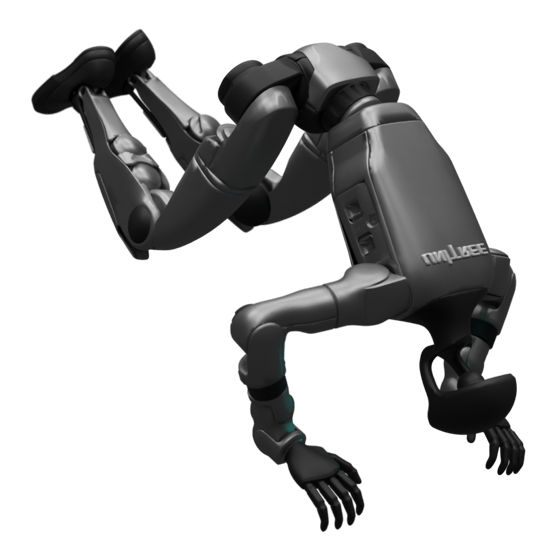}};
\node[inner sep=0] (f3) at (0.5,4.75)
    {\includegraphics[height=14mm,width=11mm,keepaspectratio]{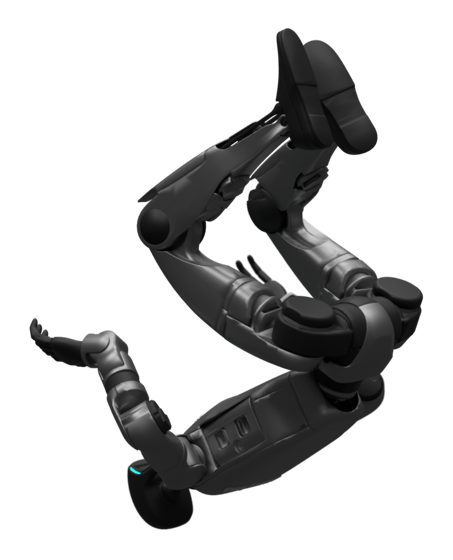}};
\node[inner sep=0] (f4) at (1.7,4.75)
    {\includegraphics[height=14mm,width=11mm,keepaspectratio]{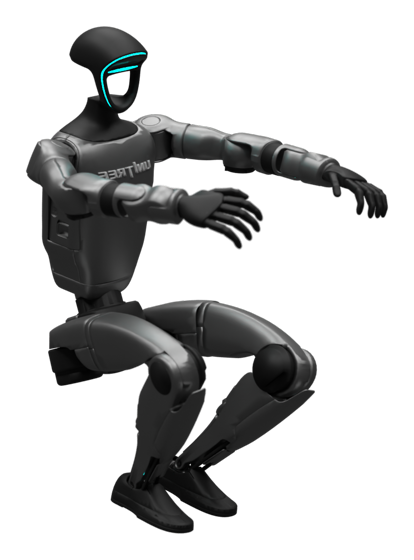}};
\node[font=\footnotesize, align=center] at (-0.1,3.8)
    {full future poses $t{+}1,\dots,t{+}k$};
\draw[tarr] ([xshift=-1mm]f4.east |- gt.west) -- (gt.west);
\draw[tarr] (gt.north) -- ++(0,0.45) -| (trk.north);
\node[inner sep=0] (cur) at (0.65,1.05)
    {\includegraphics[height=19mm]{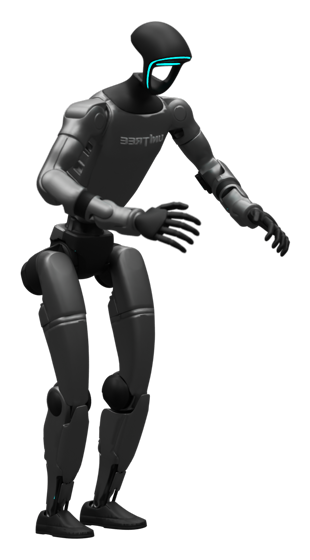}};
\node[font=\footnotesize] at (0.65,-0.1) {current pose $t$};
\draw[arr] ([xshift=-2mm]cur.east |- hist.west) -- (hist.west);

% ---------------- legend ----------------
\draw[arr] (6.0,0.45) -- (6.8,0.45)
    node[right, font=\footnotesize] {training and inference};
\draw[tarr] (6.0,0.0) -- (6.8,0.0)
    node[right, font=\footnotesize] {training only};
\end{tikzpicture}%
}
\caption{\textbf{Model and training.} The prior maps partial prompts and observation history to a deterministic condition and a latent distribution. The decoder combines the condition with a sphere-projected latent to produce joint targets. Training adds a posterior conditioned on the full reference and supervision from a frozen tracker. Only the solid path runs at deployment. Pose renders illustrate state, not camera inputs.}
%\caption{\textbf{Model and training.} The prior maps a partial prompt and observation history into a deterministic condition and a latent distribution. A latent sample is projected onto the unit sphere, fused with the condition, and decoded to joint targets. Training adds a posterior informed by the full reference and a frozen tracker that supervises the decoded actions. KL and sequential losses regularize learning. At deployment only the solid path remains. Pose renders illustrate state, not camera inputs.}
\label{fig:architecture}
\vspace{-0.2cm}
\end{figure*}

Training proceeds in two phases. Dense motion trackers learn whole-body coordination from complete reference motions using privileged simulator state (\Cref{sec:global-tracker}). Masked on-policy distillation then trains one generative policy per embodiment, conditioned on partial prompts through a flexible interface (\Cref{sec:prompt,sec:distillation}). Distillation can combine multiple specialist trackers, with each frozen expert labeling states from its own training group. The deployable policy $\pi_\theta(a_t\mid o_t,a_{t-1},g_t)$ maps a prompt $g_t$, current observations and sensor history $o_t$, and the previous action $a_{t-1}$ to joint actions using only deployment-available inputs. Once frozen, stateful programs called \emph{skills} construct prompts to express diverse behaviors without retraining (\Cref{sec:prompt-compiler}).
% Dense reference motions provide detailed training supervision, while runtime tasks may specify only sparse goals. We bridge these settings in two phases: first, a dense motion tracker learns whole-body coordination from complete reference motions using privileged simulator state (\Cref{sec:global-tracker}); second, masked on-policy distillation trains a generative policy conditioned on partial prompts (\Cref{sec:distillation}). The policy $\pi_\theta(a_t\mid o_t,a_{t-1},g_t)$ receives only deployment-available inputs and maps a prompt $g_t$, sensor history $o_t$, and the previous action $a_{t-1}$ to joint actions $a_t$. Once frozen, it serves as the execution layer introduced in \Cref{sec:introduction}: stateful programs called \emph{skills} construct prompts at each control step to express task intent---steering, recovery, object interaction---without retraining the policy (\Cref{sec:prompt-compiler}). %Embodiment-specific adaptations are described in \Cref{sec:embodiment-observations}.

\subsection{Dense Motion Tracker}
\label{sec:global-tracker}

An MLP policy learns whole-body coordination from complete reference motions:
\begin{equation}
 a_t^{\mathrm{trk}}=\pi^{\mathrm{trk}}
 \bigl(s_t^{\mathrm{trk}},y_t,a_{t-1}\bigr),
 \label{eq:tracker-policy}
\end{equation}
where $s_t^{\mathrm{trk}}$ includes privileged simulator state and $y_t$ encodes complete reference states at a configured set of future offsets. We optimize the tracker with PPO~\cite{schulman2017ppo}. Its reward combines tracking terms~\cite{peng2018deepmimic} and action-rate regularization~\cite{liao2026beyondmimic}:
\begin{equation}
 r_t=\sum_k w_k\exp(-\lambda_k e_{k,t})
      -w_a\mathcal R_{\mathrm{act},t}+r_t^{\mathrm{extra}}.
 \label{eq:tracker-reward}
\end{equation}
The errors $e_{k,t}$ measure body position, orientation, and velocity tracking; $w_k$ and $\lambda_k$ set their weights and sensitivities. The action regularizer $\mathcal R_{\mathrm{act},t}$ penalizes changes between consecutive commands, and $r_t^{\mathrm{extra}}$ collects selected-body precision terms and embodiment-specific penalties. These rewards train the trackers, not the distilled student. The supplementary material details the G1 BONES and SOMA experts' rewards, architectures, and optimization settings.

\subsection{Prompt Interface}
\label{sec:prompt}

To support partial goals at runtime, we distill the frozen trackers into \method. At timestep $t$, a prompt combines four optional goal streams and scene context,
\begin{equation}
 g_t=\left(\mathcal C_t^w,\mathcal C_t^\ell,u_t,\xi_t,b_t\right).
 \label{eq:prompt-contract}
\end{equation}
Masks identify which bodies, properties, and timesteps are constrained, so an omitted goal can remain unspecified.

\smallskip
\noindent\textbf{Spatio-temporal goals:} $\mathcal C_t^w$ specifies body positions or orientations in world coordinates, each with a time-to-arrival. Position and orientation constraints are independently optional: a prompt can request a planar location while leaving height free, specify a timed hand target, or supply dense whole-body poses.

\smallskip
\noindent\textbf{Canonicalized pose goals:} $\mathcal C_t^\ell$ describes the desired pose independently of its world placement. Horizontal positions are relative to the target torso and aligned with its heading; heights are measured above the target pose's lowest body point. A raised-hand pose therefore retains its meaning regardless of world position or heading. These goals are not offsets from the robot's current pose.

\smallskip
\noindent\textbf{Velocity:} $u_t$ specifies planar travel velocity in the agent's heading frame, without specifying an endpoint.

\smallskip
\noindent\textbf{Style cue:} $\xi_t$ specifies the character of the motion through an example clip. A small set of frames provides canonicalized body poses and optional heading-aligned velocities. Pooling these frames into one token discards temporal order: the cue describes example motion states without assigning arrival times. A location goal can therefore be paired with running or crouched-walking examples without prescribing their trajectory.

\smallskip
\noindent\textbf{Scenes:} For object interaction, each scene element is decomposed offline into bounding boxes $b_t$ (e.g.\ a sofa yields seat and backrest boxes). The policy observes the eight corner positions of each box in the agent's heading frame, from which both position and orientation are recoverable. This compact, fixed-width representation needs only pre-measured dimensions and an object-pose estimate at deployment.

\subsection{Generative Policy}
\label{sec:distillation}

A partial prompt admits many valid motions. Regressing a single action directly from that prompt can average incompatible motion completions~\cite{tessler2024maskedmimic}. We instead use a conditional variational autoencoder (VAE): a privileged posterior uses the full reference during training, while a prior uses only deployment-available observations and the prompt. A deterministic condition $d_t$ encodes the observed context; a stochastic latent selects a completion. We project this latent onto the unit sphere~\cite{xu2026interprior} before combining it with $d_t$. With $x_t=(o_t,a_{t-1})$, privileged simulator state $s_t$, and complete reference $y_t$, the distributions before projection are
\begin{equation}
 \begin{aligned}
 p_\theta(\tilde z_t\mid x_t,g_t)
   &=\mathcal N(\mu_t,\operatorname{diag}(\sigma_t^2)),\\
 q_\phi(\tilde z_t\mid x_t,g_t,s_t,y_t)
   &=\mathcal N(\mu_t+\Delta\mu_t,\operatorname{diag}({\sigma_t'}^2)).
 \end{aligned}
 \label{eq:prior-posterior}
\end{equation}
The prior transformer combines sensor and prompt tokens to predict $d_t$, $\mu_t$, and $\sigma_t$; masks exclude unavailable goals (\Cref{fig:architecture}). A separate training-only posterior encoder predicts the mean correction $\Delta\mu_t$ and its own variance from privileged state, reference, and prompt features. For either distribution, a sample is projected as $z_t=\tilde z_t/\|\tilde z_t\|_2$ and fused with $d_t$ as $h_t=f_\theta(d_t,z_t)$. The decoder maps the current observations, previous action, and $h_t$ to an action. Superscripts $p$ and $q$ denote quantities computed with prior and posterior samples, respectively. A training-only predictor $F$ estimates the next condition from $h_t^q$, and a temporal consistency loss smooths consecutive fused representations~\cite{schwarzer2021spr,hansen2024tdmpc2}.

\subsection{Training}
\label{sec:training}

Distillation collects states under a mixture of posterior-guided student and frozen-tracker rollouts. The tracker labels each visited state using its complete reference, while the student receives randomly masked prompts. The student minimizes
\begin{equation}
 \begin{aligned}
 \mathcal L&={}\mathbb E\bigl[
   \ell(a_t^q,a_t^{\mathrm{trk}})
   +\beta_e D_{\mathrm{KL}}(q_\phi\|p_\theta)\bigr] + \lambda_\mu\mathbb E(\|\mu_t\|_2-1)^2\\
 &+\lambda_{\mathrm{dyn}}\mathbb E
   \bigl[\ell(F(h_t^q),\operatorname{sg}(d_{t+1}))\bigr] + \lambda_{\mathrm{tc}}\mathbb E\bigl[
   \ell(h_t^q,\operatorname{sg}(h_{t-1}^q))\bigr]\\
 &+\lambda_{\mathrm{act}}\mathbb E\bigl[
   \ell(a_t^q,\operatorname{sg}(a_{t-1}^q))\bigr],
 \end{aligned}
 \label{eq:distillation-loss}
\end{equation}
where $a_t^q$ is the posterior-decoded student action, $\ell$ averages squared error over coordinates, and $\operatorname{sg}$ treats the target as a constant during optimization. The first term imitates the frozen tracker; KL aligns the prior and posterior Gaussians before projection. The remaining terms encourage a unit-norm prior mean, predict the next condition, smooth consecutive representations, and smooth consecutive student actions. Temporal terms use only within-episode pairs, excluding resets. The schedule $\beta_e$ balances reconstruction and prior--posterior alignment. For G1, cyclical annealing~\cite{fu2019cyclical} repeatedly relaxes then strengthens this constraint; SOMA uses a single ramp followed by a constant weight. The supplementary material gives the schedules, reductions, coefficients, and SOMA's additional diversity loss. At deployment, only the prior, fusion module, and decoder are needed.

\section{\method: Skill Layer}
\label{sec:prompt-compiler}

An underspecified prompt can produce several valid motions, some less stable or desirable than others. A velocity command alone may yield foot stutter or hesitant steps. A \emph{skill} is a stateful program that constructs and updates prompts at each control step, supplementing high-level intent with additional goals that guide the policy toward the desired behavior. The \emph{harness} runs the active skill and manages transitions between skills.

Because the policy tolerates approximate goals (\Cref{sec:method}), a skill does not need to plan precise trajectories. Instead, it draws a \textbf{rough sketch of the desired motion}: a few geometric targets such as static points, straight lines, or simple periodic curves. The policy renders this sketch into coordinated whole-body movement, handling balance, weight transfer, and joint actions. A sketch with only a handful of parameters suffices; a separately trained motion planner is not required. %Overfitting is structurally difficult when the rule has so few degrees of freedom.
%%JP: not sure if this last sentence about overfitting is really necessary.

\begin{figure}[!t]
    \centering
    \includegraphics[trim={0.1cm 0.25cm 0.1cm 0.25cm},clip,width=0.99\linewidth]{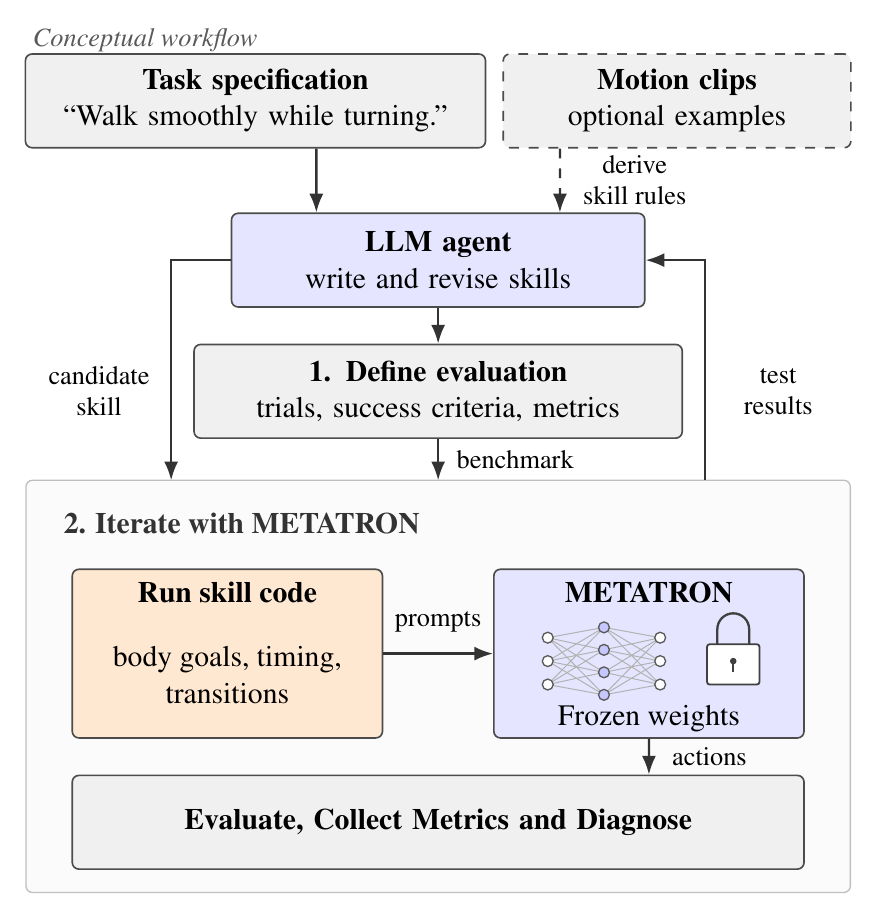}
    \caption{\textbf{Agent-assisted skill development.} Given a task specification and optional motion examples, a human and LLM agent define evaluation scenarios and criteria. The LLM agent then iteratively refines the skill code using simulation feedback. Skills specify goals for the frozen \method policy, enabling the model to perform new tasks without additional training.}
    \label{fig:skill-workflow}
    \vspace{-0.2cm}
\end{figure}

Authoring a skill follows a common pattern (\Cref{fig:skill-workflow}). A human writes a first-pass rule describing goals required by the task. For simple cases such as compliance or recovery, this rule is already sufficient. When visible artifacts remain, such as inefficient stepping during locomotion or hesitant weight shifts when rising from a chair, the human describes these issues and a coding agent takes over refinement. The agent constructs an evaluation playground that spans the relevant variation: different furniture geometries and approach angles for sitting, or randomized velocity sequences for steering. It then analyzes motion-capture clips and fits compact rules that capture shared characteristics across the corpus. The agent scores each candidate on the playground, identifies failing cases, and iterates.

We showcase example skills constructed by the agent below:

% A steering command specifies a target travel velocity and heading, leaving the stepping pattern unspecified. We used Claude Code to fit a periodic foot-trajectory rule from locomotion clips in the motion corpus. The entire rule is a sinusoidal gait with three constants---cadence, swing amplitude, and stance width---each a linear function of speed. 
% %%JP: if the rule is simple, it might be helpful to write out the equation for it so readers can better understand what the agent came up with
% The angle between travel direction and the robot's facing determines the swing axis and widens the stance for lateral movement. The skill evaluates these trajectories at future arrival times to construct foot-placement goals, alongside torso targets that follow the requested velocity and heading. The skill specifies the gait pattern; the policy specifies whole-body coordination and joint actions.

\smallskip
\noindent\textbf{Steering:}
A steering command specifies velocity and heading but leaves stepping unspecified. With only these constraints, the policy sometimes produces small, high-frequency steps that waste energy and transfer unsafely to hardware. We used coding agents to fit a single periodic foot-trajectory rule from locomotion clips in a benchmark of randomized velocity sequences. The rule is governed by two quantities: travel speed $v$ and the angle $\alpha$ between travel direction and the robot's facing. At future time $\tau$, the foot target in the planned torso's heading frame is
\begin{equation}
    \mathbf{p}_i(\tau)=\mathbf{b}_i+
    \begin{bmatrix}
        A(v)\cos\alpha\cos\phi_i\\
        W\bigl(s_i|\sin\alpha|+|\cos\alpha|\sin\alpha\cos\phi_i\bigr)\\
        H\max(0,-\sin\phi_i)
    \end{bmatrix},
\end{equation}
where $\mathbf{b}_i$ is the calibrated stance offset, $s_R=-1$, $s_L=1$, and phase $\phi_i=\phi_t+2\pi f(v)\tau+\delta_i$ alternates between feet ($\delta_R=0$, $\delta_L=\pi$). The fitted functions $A(v)$ and $f(v)$ set swing amplitude and cadence; $W$ and $H$ control lateral widening and foot lift. Each foot swings with the travel command, lifts at mid-swing, and returns to a stance that widens as travel turns lateral. Cadence and step length both grow with speed, producing longer, quicker strides rather than faster shuffling. Forward walking, side-stepping, and turning emerge from this rule at different values of $\alpha$. Fitted coefficients and deployment details are provided in the supplementary material.

\smallskip
\noindent\textbf{Sitting:}
Sitting on furniture requires only coarse goals derived from the seat's bounding-box geometry. A coding agent fit the goal rule from a benchmark of seated interactions. The result is three static target points: a torso target at a fixed height above the seat surface, slightly behind the seat center, and both feet just past the front edge at stance width. All prompt slots receive these three targets, without intermediate waypoints. The policy coordinates turning and stepping before sitting as needed. Standing reverses the pattern: the skill raises the torso target along a straight line from seated to standing height while marching it continuously forward. No stepping pattern is prescribed, yet the policy produces coordinated steps to rise and walk away.

\smallskip
\noindent\textbf{Recovery:}
Recovery returns the robot from a fallen configuration to an upright pose.
A hand-authored rule is sufficient: the simulation skill specifies a canonicalized upright pose, while the G1 deployment skill constrains only torso height. Arrival time scales with the remaining height deficit. A lower entry threshold and higher exit threshold prevent repeated switching near the boundary. Automatic recovery is disabled during dense tracking and unconditional generation; the supplementary material gives the thresholds and target construction.

% Required in the preamble:
% \usepackage{graphicx}
% \usepackage{tikz}
% \usetikzlibrary{arrows.meta}

\begin{figure}[t]
\centering
% Single-column drawing at final size; no global scaling of labels.
\begin{tikzpicture}[
    font=\fontsize{8.5}{10}\selectfont,
    card/.style={draw=black!60, rounded corners=2pt, line width=0.5pt,
        minimum width=31mm, minimum height=31mm, anchor=south west},
    arr/.style={-{Latex[length=1.7mm]}, line width=0.65pt, draw=black!75},
    thin/.style={line width=0.45pt, draw=black!65},
    guard/.style={fill=white, inner sep=1.5pt},
    title/.style={font=\fontsize{9}{10.5}\selectfont\bfseries}
]
\node[card,fill=orange!8] (steer) at (0.05,0) {};
\node[card,fill=green!7] (rec) at (5.2,0) {};
\node[title] at (1.6,2.86) {Steer};
\node[title] at (6.75,2.86) {Get up};

% Timed torso and anti-phase foot targets, viewed from above.
\draw[thin,line width=0.8pt] (0.65,2.10) -- (0.65,2.50);
\fill[black!75] (0.65,2.54) circle (2.2pt);
\draw[thin,line width=0.8pt] (0.42,2.10) -- (0.88,2.10);
\draw[arr,draw=blue!55!black] (1.1,2.30) -- (2.35,2.30)
    node[right,text=blue!55!black] {move};
\fill[black!75] (0.45,1.55) circle (2.2pt);
\draw[thin,dashed,draw=black!40] (0.45,1.55) -- (2.85,1.55);
\foreach \x in {0.85,1.32,1.79,2.26,2.73}{
    \draw[thin,fill=white] (\x,1.55) circle (1.7pt);
    \draw[thin] (\x,1.55) -- ++(0.12,0);
}
\foreach \x/\y in {0.65/1.90,1.12/1.20,1.59/1.90,2.06/1.20,2.53/1.90}{
    \draw[thin,fill=white] (\x,\y) ellipse (3.3pt and 1.7pt);
}
\node[align=center] at (1.6,0.40)
    {torso and foot targets\\at future arrival times};

% Recovery requests an upright posture rather than a joint-space trajectory.
\draw[thin] (5.4,0.75) -- (8.08,0.75);
\node[inner sep=0,anchor=south] at (5.97,0.75)
    {\includegraphics[height=10mm]{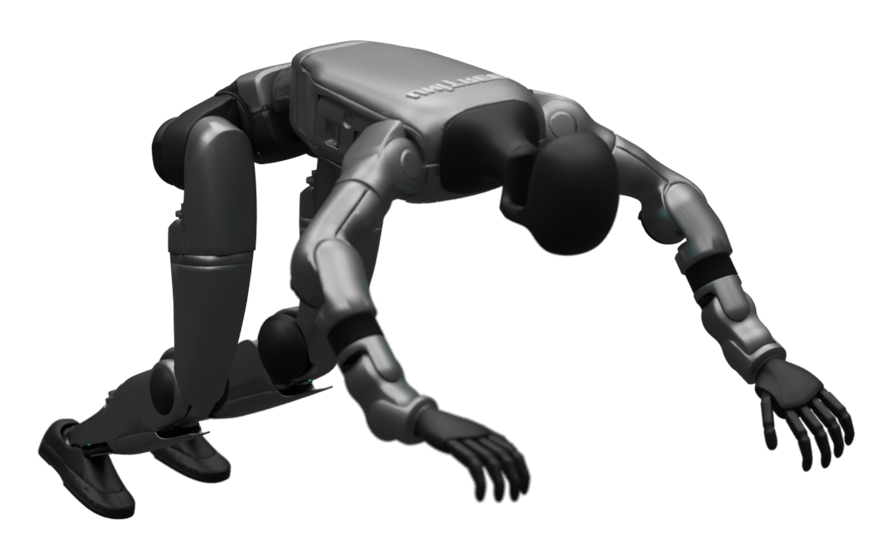}};
\node[inner sep=0,anchor=south,opacity=0.6] at (7.45,0.75)
    {\includegraphics[height=17mm]{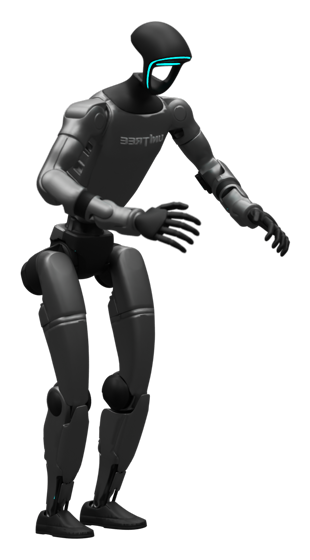}};
\draw[arr,dashed]
    (5.97,1.83) -- (5.97,2.60)
    -- node[guard,below,align=center] {rise\\time}
    (7.45,2.60) -- (7.45,2.48);
\node[align=center,text=green!35!black] at (6.75,0.35)
    {upright target pose\\all bodies};

% Hysteretic switching changes only the prompt.
\draw[arr,draw=red!55!black]
    ([yshift=4.5mm]steer.east) --
    node[guard,above] {fall detected}
    ([yshift=4.5mm]rec.west);
\draw[arr,draw=green!45!black]
    ([yshift=-4.5mm]rec.west) --
    node[guard,below] {upright again}
    ([yshift=-4.5mm]steer.east);

% Bring the shared policy closer to the panels.
\node[draw=black!60,rounded corners=2pt,fill=blue!9,align=center,
    minimum width=40mm,minimum height=8mm,line width=0.5pt]
    (pol) at (4.175,-0.55)
    {selected prompt\\one frozen policy};
\draw[arr] (steer.south) |- (pol.west);
\draw[arr] (rec.south) |- (pol.east);
\end{tikzpicture}
\caption{\textbf{Change the prompt, retain the controller.}
In this simulation skill, a detected fall switches steering goals to a whole-body upright pose; steering resumes once upright. G1 deployment instead requests torso height alone.}
\label{fig:recovery-fsm}
\vspace{-0.2cm}
\end{figure}

\smallskip
\noindent\textbf{Interactive compliance:}
Physical human-robot interaction requires maintaining stance while accommodating human-guided arm motion. A hand-authored rule is again sufficient. The skill commands zero travel velocity and constrains torso height and heading, leaving arm poses unspecified. Unlike full-pose tracking, these prompts do not require the arms to return to a prescribed pose, so the policy treats externally induced motion as consistent with the prompt. In our demonstrations, this produces handshakes and guided manipulation without a separate compliance controller.

These skills re-purpose existing behaviors into new capabilities without requiring any skill-specific training.

\section{Training Data}
\label{sec:training-data}

For each embodiment, motion-capture data, scene interactions, and synthetic recoveries provide supervision for the distilled policy (\cref{sec:distillation}). G1 partitions these sources among four specialist trackers (\cref{sec:global-tracker}): BONES, AMASS+LAFAN, recovery, and SAMP. Kinematic retargeting adapts motions to the G1 morphology; scene interactions additionally require collision handling, while recoveries are generated directly for each embodiment \cite{kim2025pyroki}.

\smallskip
\noindent\textbf{Motion-Capture Retargeting:} Our base dataset is a large-scale motion-capture library~\cite{bonesstudio} spanning locomotion, dance, acrobatics, sports, and everyday activities.

\smallskip
\noindent\textbf{Scene-Interaction Retargeting:} \label{sec:samp-retargeting} Scene interactions require preserving body--object contacts as well as the motion. We use SAMP~\cite{hassan2021samp}, which provides human motions such as sitting on furniture and lying on beds, represented with the SMPL body model. Kinematic retargeting alone can introduce furniture collisions or distort the posture to avoid them.

We use two-phase optimization. The first phase optimizes keypoint alignment, joint limits, smoothness, and foot--ground contact without scene geometry. The second adds a scene-collision cost and a joint-anchor penalty, favoring root translation over changes to the joint configuration. We represent furniture with an oriented surface point cloud, filtering interior surfaces per motion. The cost penalizes penetration depth but not contact, allowing contact without hovering. We will release the retargeted SAMP dataset upon publication.
%Furniture is represented by an oriented surface point cloud, with interior surfaces filtered per motion to avoid spurious repulsion. The collision cost is zero at contact and increases with penetration depth, allowing the body to rest on the furniture without hovering. We will release the retargeted SAMP dataset upon publication.

\smallskip
\noindent\textbf{Synthetic Recovery Motions:} \label{sec:recovery-data} Recovery motions must accommodate the embodiment's dynamics and ground contacts. We therefore generate them directly using each simulated embodiment. We extend Hybrid Imitation Learning (HIL)~\cite{wang2026hil} with MaskedMimic-style masking~\cite{tessler2024maskedmimic} of body targets and arrival times. We use this policy to record recoveries from randomized fallen poses. Collection and filtering settings are given in the supplementary material.

\section{Experiments}
\label{sec:experiments}

\subsection{Experimental Setup}
\label{sec:experimental setup}

Our implementation builds on ProtoMotions~\cite{tessler2025protomotions}. We train separate \method policies for SOMA and G1. The G1 policy distills four specialist trackers covering BONES, AMASS+LAFAN, recovery, and SAMP; the SOMA policy distills a scene-aware tracker.

\smallskip
\noindent\textbf{Training:}
Trackers are trained before distillation, and skills use the resulting frozen \method policy without retraining. The G1 demonstrations use the four-expert policy. The reference-pipeline comparisons use a separate policy trained without synthetic recovery data; the ablation models are trained separately under a matched budget and parameter counts. The supplementary material lists hardware, training budgets, and hyperparameters.
%%JP: what does coherent mean in this context?
%%JP: I don't think "convergence" is the most important part. What do we hope to improve with more training? Are we trying to improve performance, robustness, consistency, or motion quality?

% Two columns by three rows keeps the montage short enough to float near its discussion.
\begin{figure*}[!t]
    \centering
    \begin{subfigure}[b]{0.49\linewidth}
       \centering
       \includegraphics[trim={8cm 20cm 10cm 25cm},clip,width=0.248\textwidth]{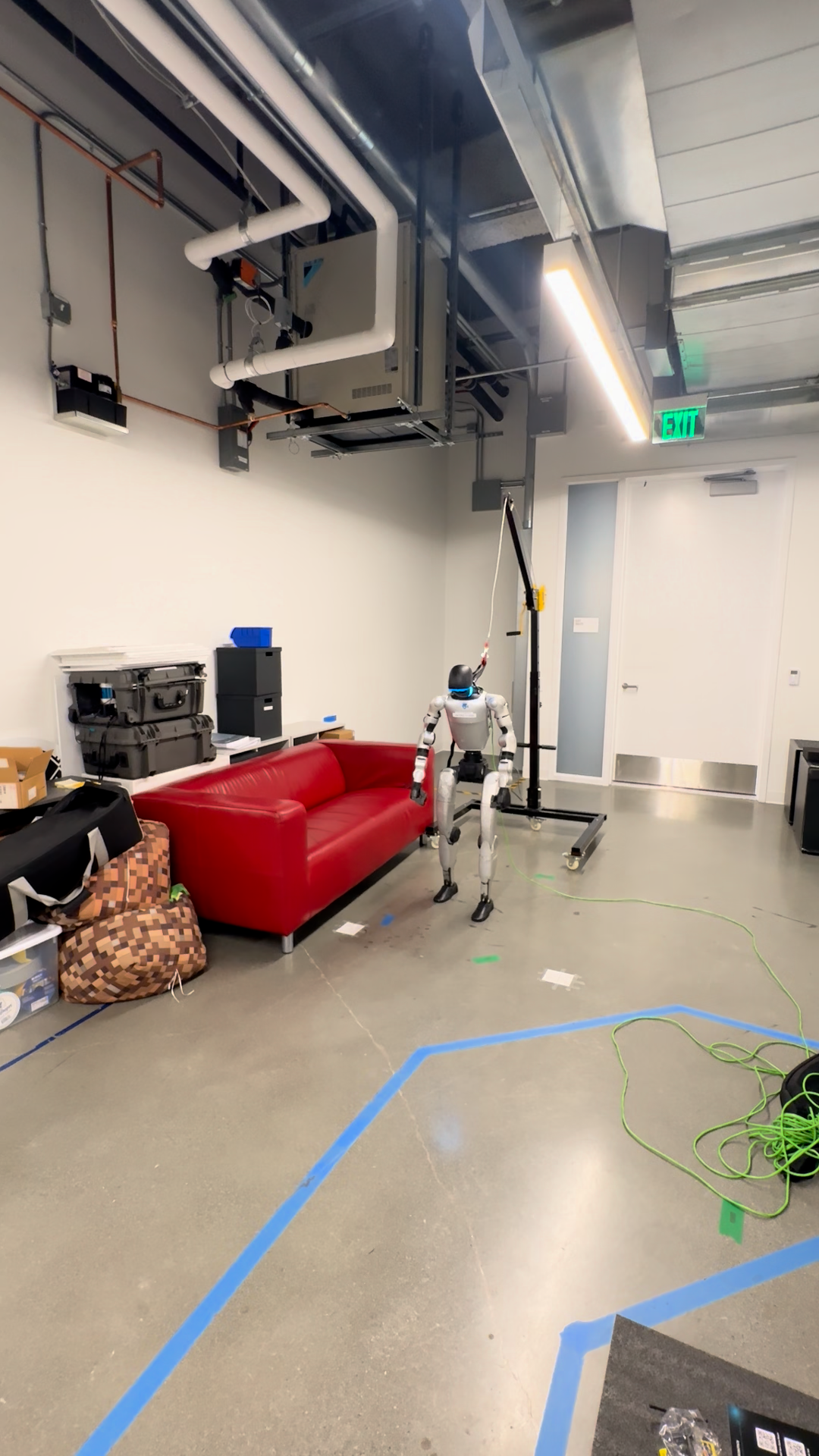}\hfill
       \includegraphics[trim={8cm 20cm 10cm 25cm},clip,width=0.248\textwidth]{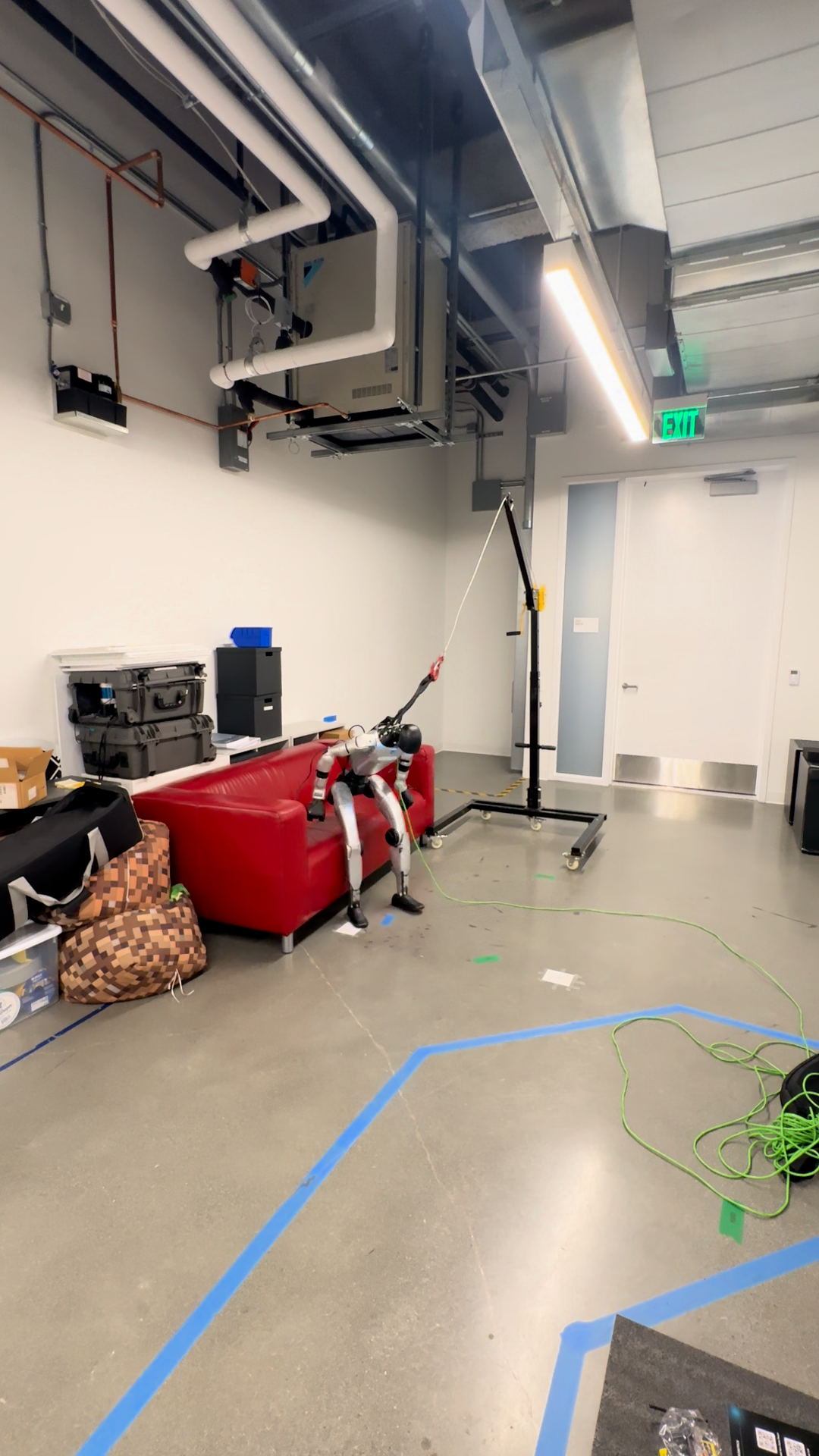}\hfill
       \includegraphics[trim={13cm 25.2cm 16cm 32.25cm},clip,width=0.248\textwidth]{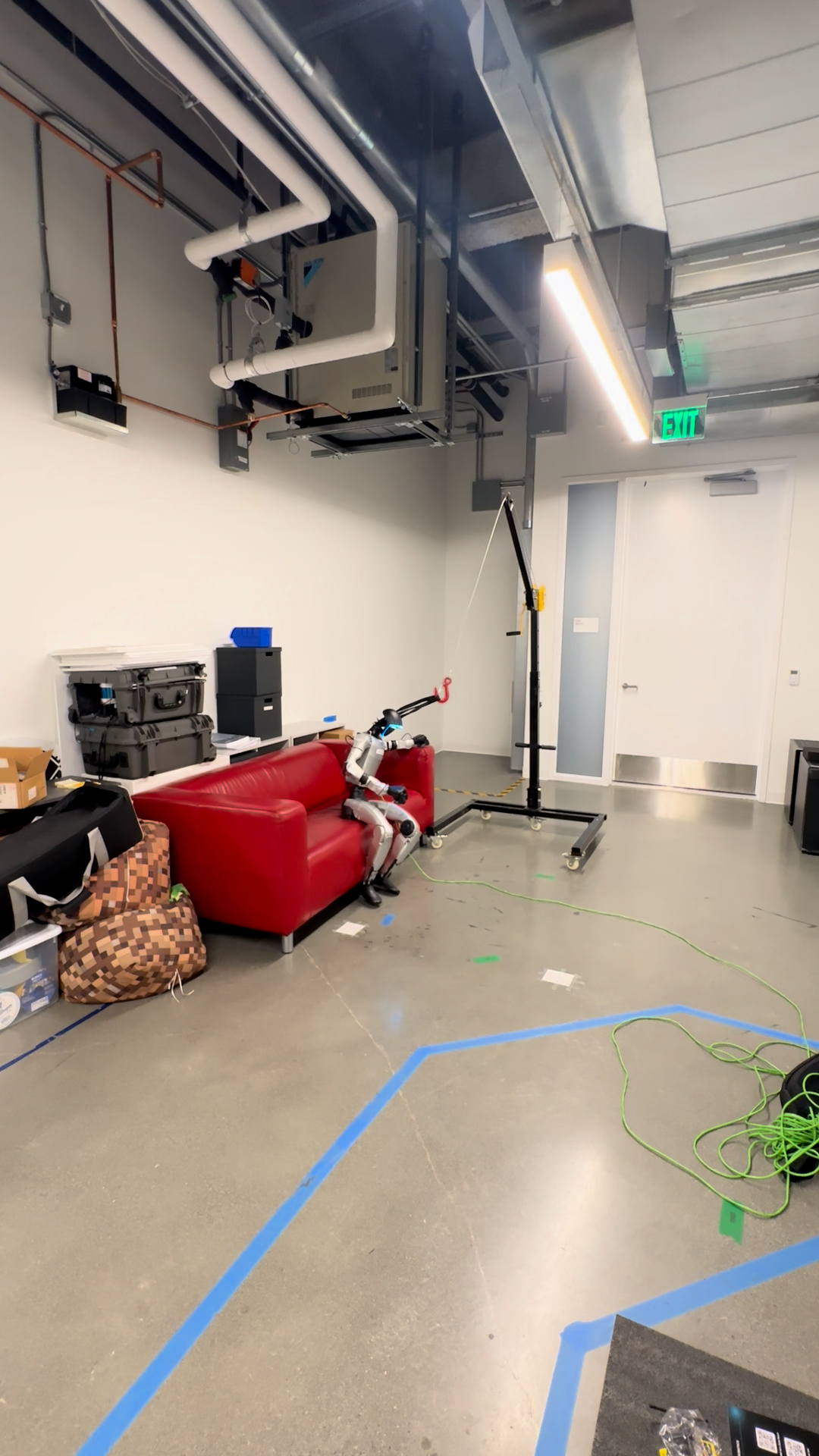}\hfill
       \includegraphics[trim={18cm 20cm 0cm 25cm},clip,width=0.248\textwidth]{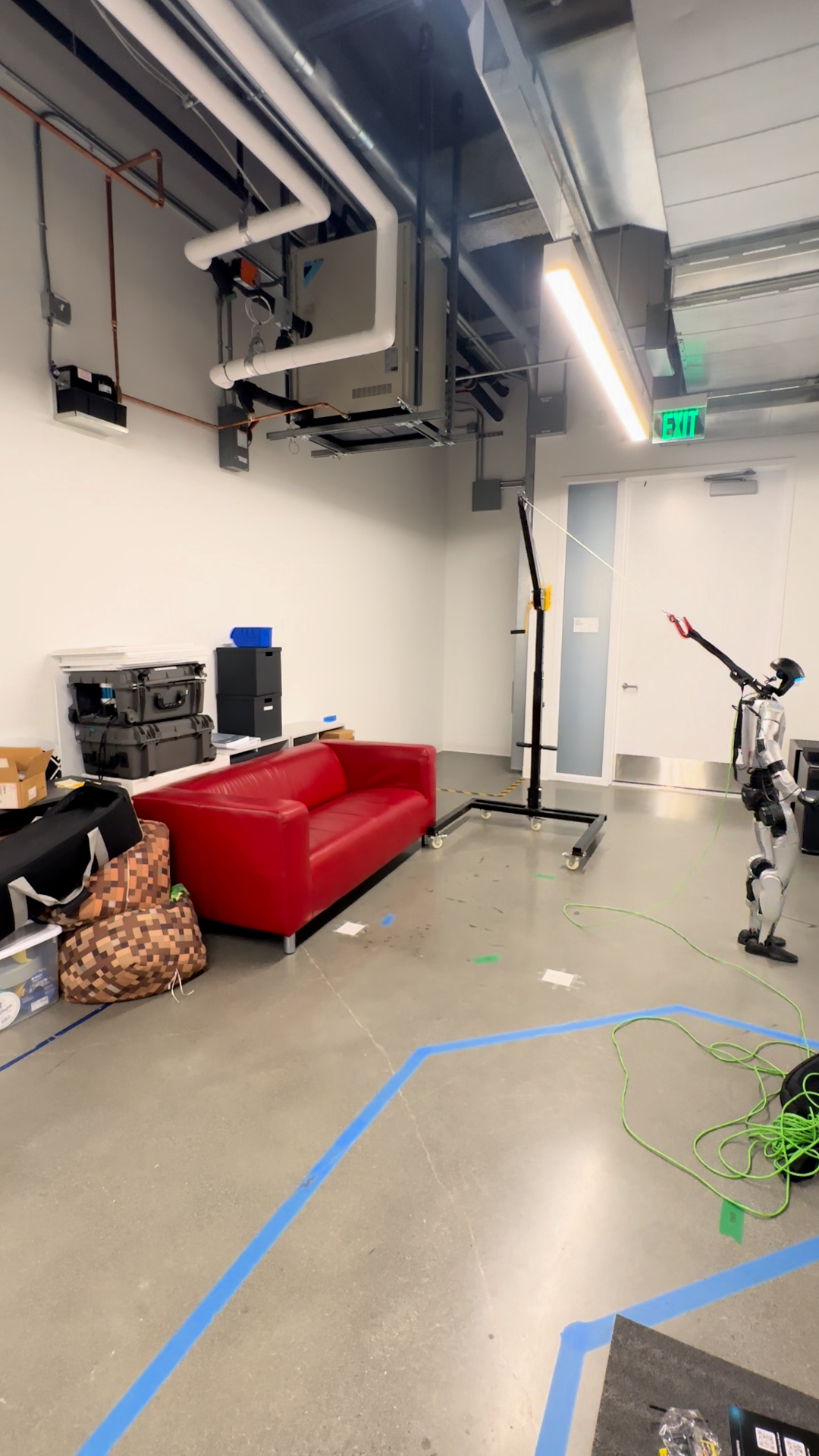}
       \textbf{Scene interaction:} G1 turns and sits on a sofa, then stands up and walks away.
   \end{subfigure}\hfill
   \begin{subfigure}[b]{0.49\linewidth}
       \centering
       \includegraphics[trim={0cm 6cm 0cm 12cm},clip,width=0.248\textwidth]{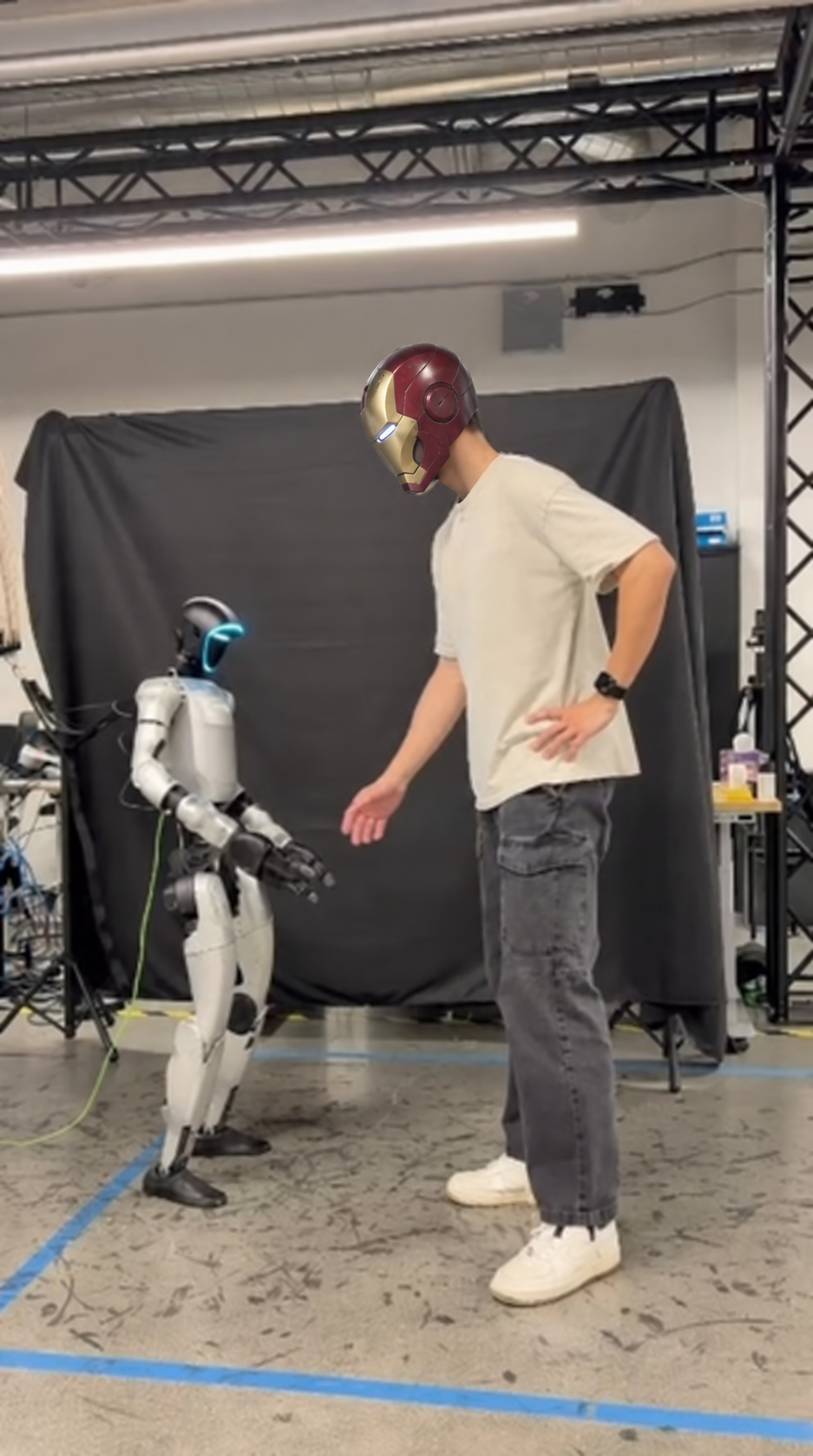}\hfill
       \includegraphics[trim={0cm 6cm 0cm 12cm},clip,width=0.248\textwidth]{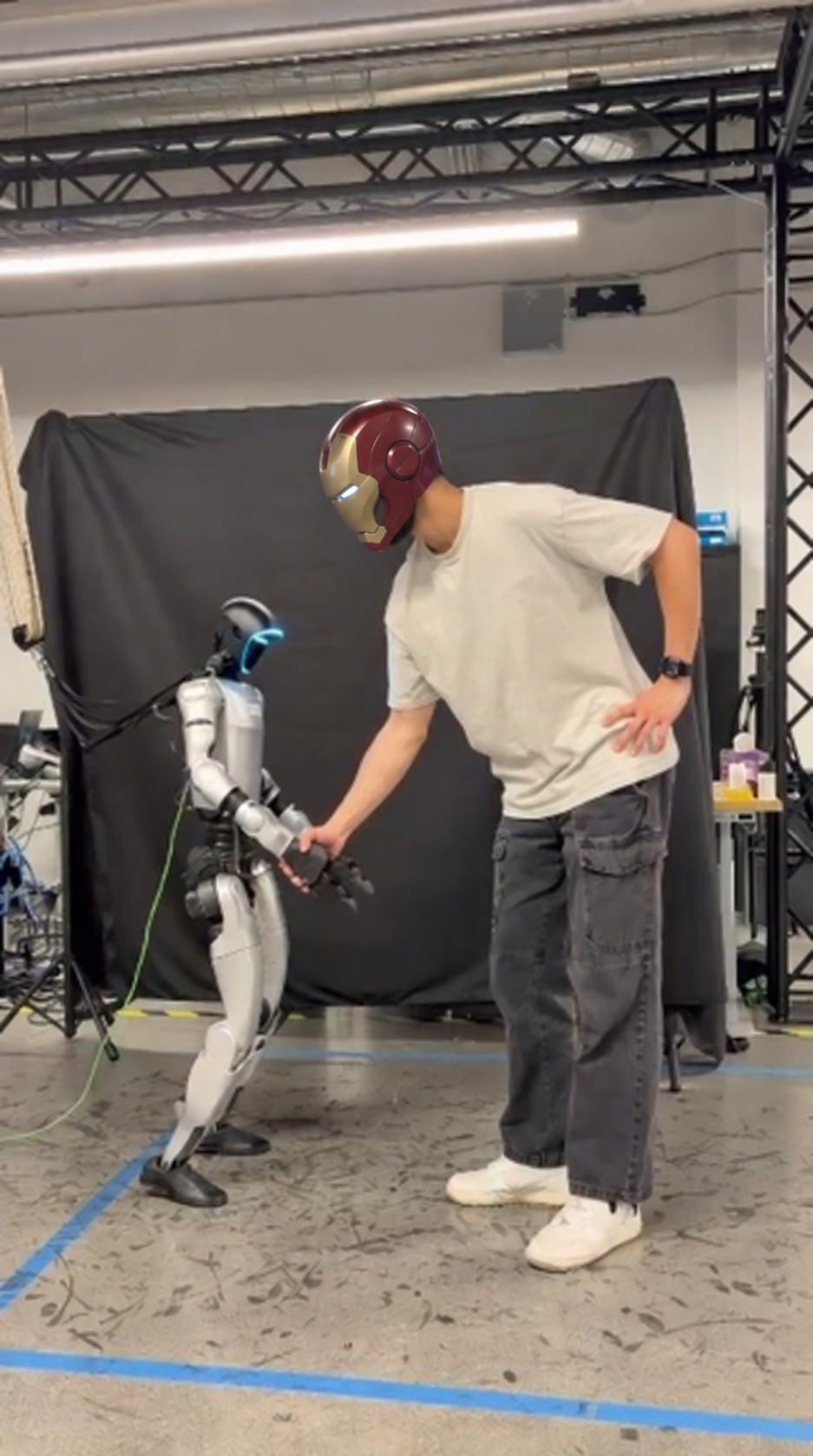}\hfill
       \includegraphics[trim={0cm 6cm 0cm 12cm},clip,width=0.248\textwidth]{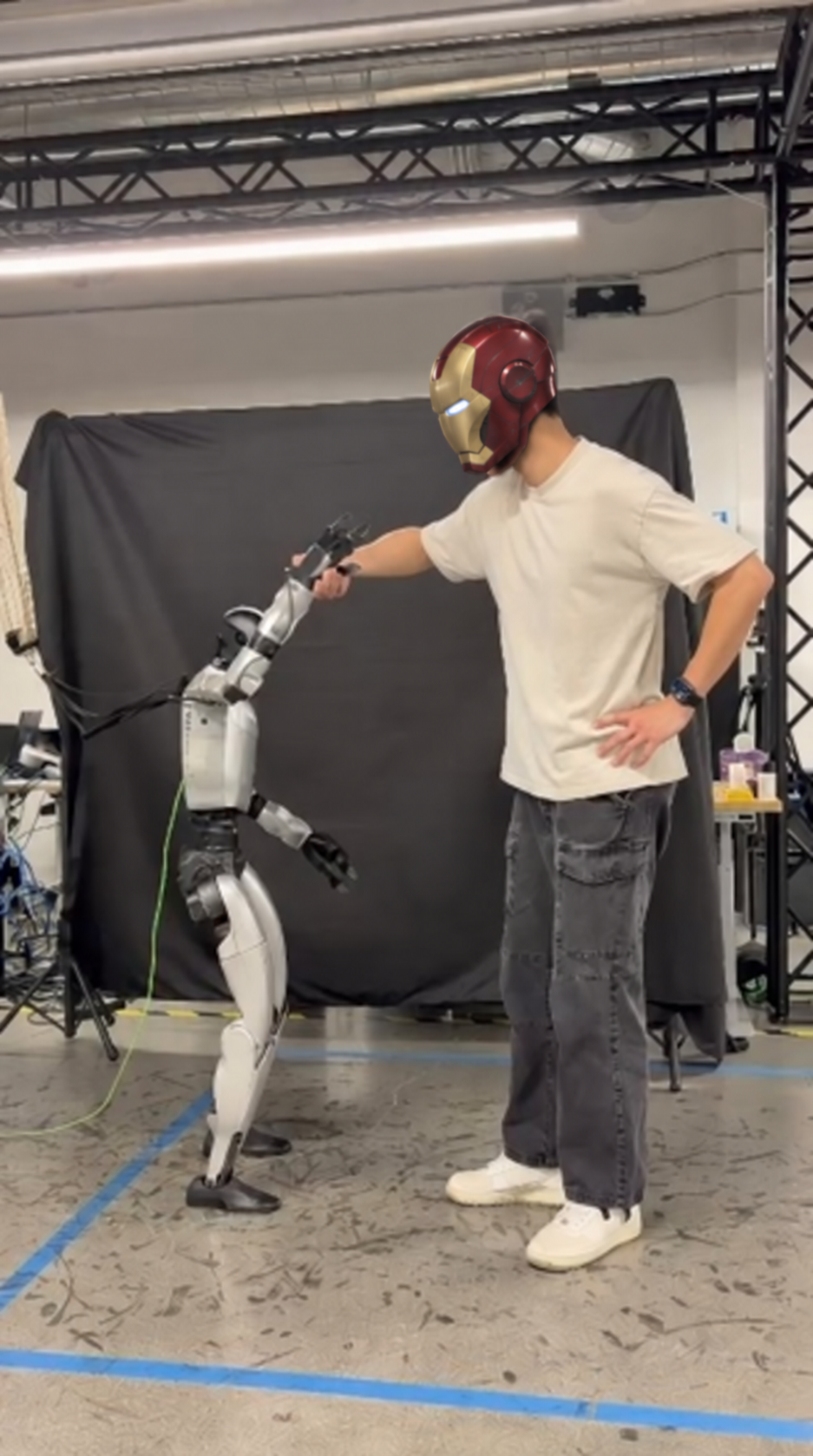}\hfill
       \includegraphics[trim={0cm 6cm 0cm 12cm},clip,width=0.248\textwidth]{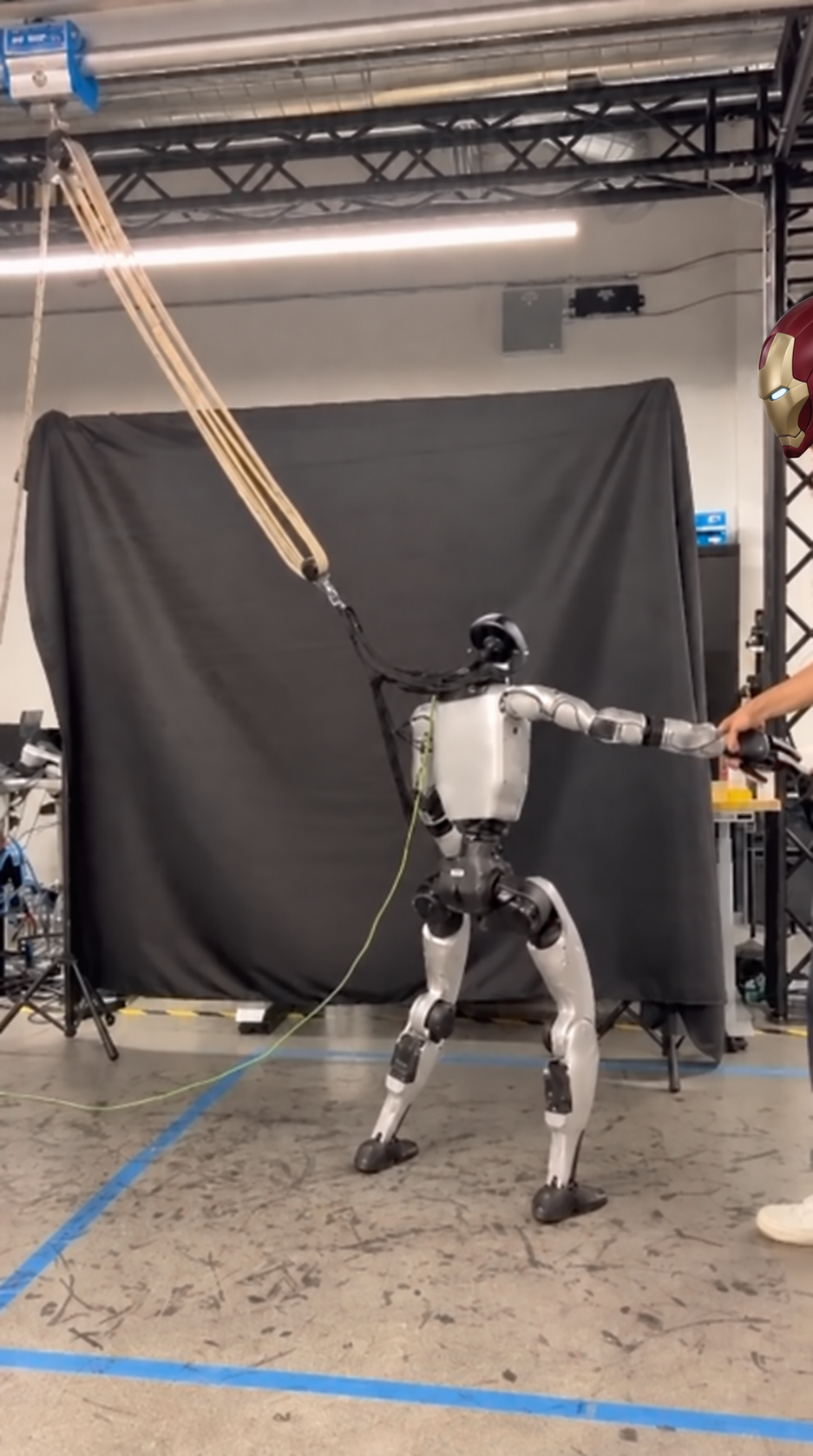}
   \textbf{Compliance:} Temporally sparse goals allow guidance.
   \end{subfigure}\\\vspace{0.2cm}
   \begin{subfigure}[b]{0.49\linewidth}
       \centering
       \reflectbox{\includegraphics[trim={0cm 2cm 0cm 2cm},clip,width=0.248\textwidth]{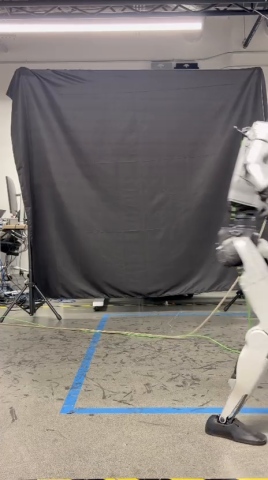}}\hfill
       \reflectbox{\includegraphics[trim={0cm 2cm 0cm 2cm},clip,width=0.248\textwidth]{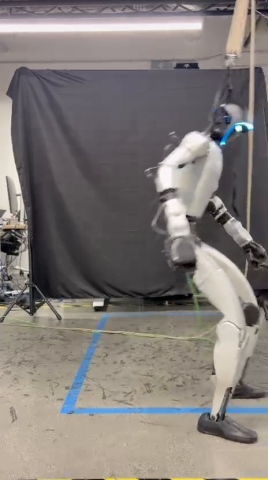}}\hfill
       \reflectbox{\includegraphics[trim={0cm 2cm 0cm 2cm},clip,width=0.248\textwidth]{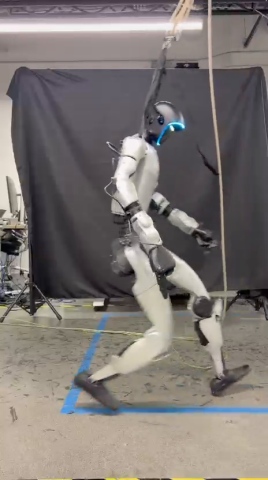}}\hfill
       \reflectbox{\includegraphics[trim={0cm 2cm 0cm 2cm},clip,width=0.248\textwidth]{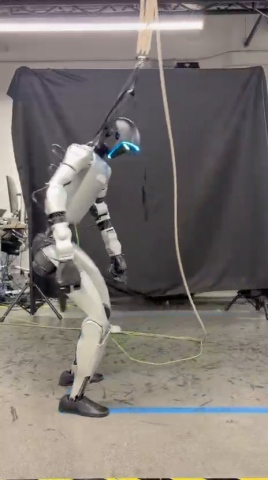}}
   \textbf{Walking backwards (steering):} An Xbox controller specifies heading and travel velocity.
   \end{subfigure}\hfill
   \begin{subfigure}[b]{0.49\linewidth}
       \centering
       \includegraphics[trim={0cm 3.5cm 5cm 6.5cm},clip,width=0.248\textwidth]{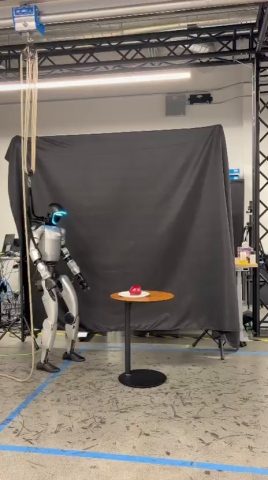}\hfill
       \includegraphics[trim={2cm 3.5cm 3cm 6.5cm},clip,width=0.248\textwidth]{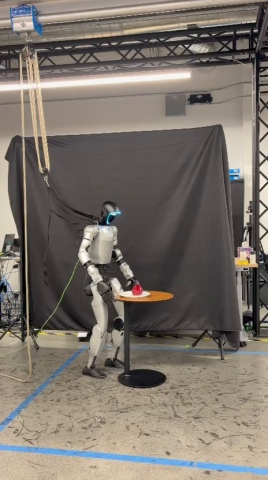}\hfill
       \includegraphics[trim={2cm 3.5cm 3cm 6.5cm},clip,width=0.248\textwidth]{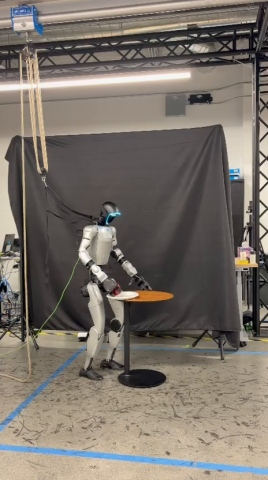}\hfill
       \includegraphics[trim={3cm 6.57cm 4cm 6.54cm},clip,width=0.248\textwidth]{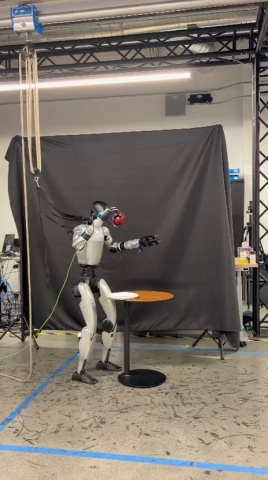}
       \textbf{Teleoperation:} VR headset provides teleoperation goals.
   \end{subfigure}\\\vspace{0.2cm}
   \begin{subfigure}[b]{0.49\linewidth}
       \centering
       \includegraphics[trim={2cm 11cm 3cm 8cm},clip,width=0.992\textwidth]{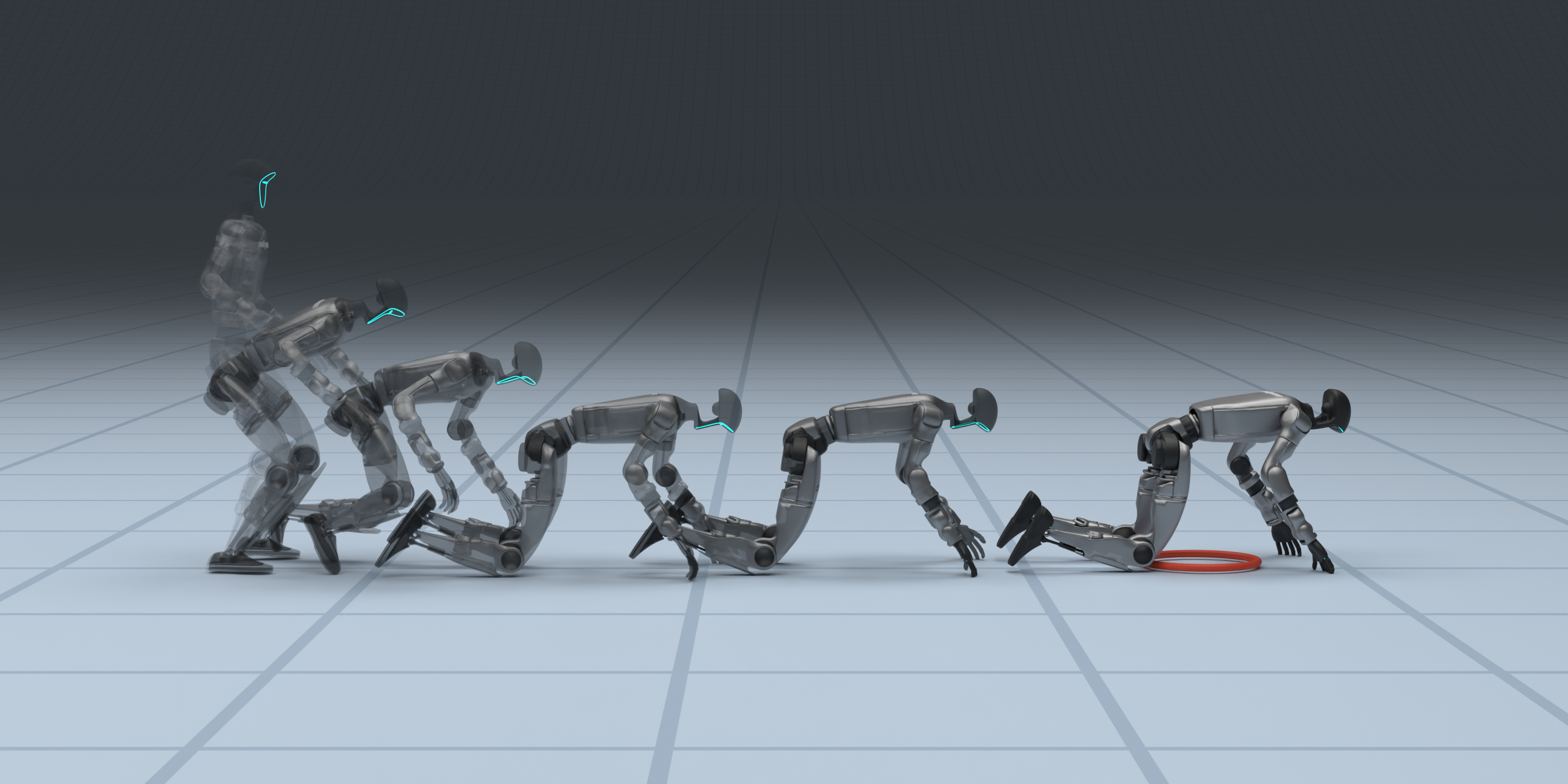}
       \textbf{Style control (simulation):} An $[x,y]$ goal and five static crawling poses guide goal reaching.
   \end{subfigure}\hfill
  \begin{subfigure}[b]{0.49\linewidth}
       \centering
       \includegraphics[trim={22cm 5cm 22cm 3cm},clip,width=0.248\textwidth]{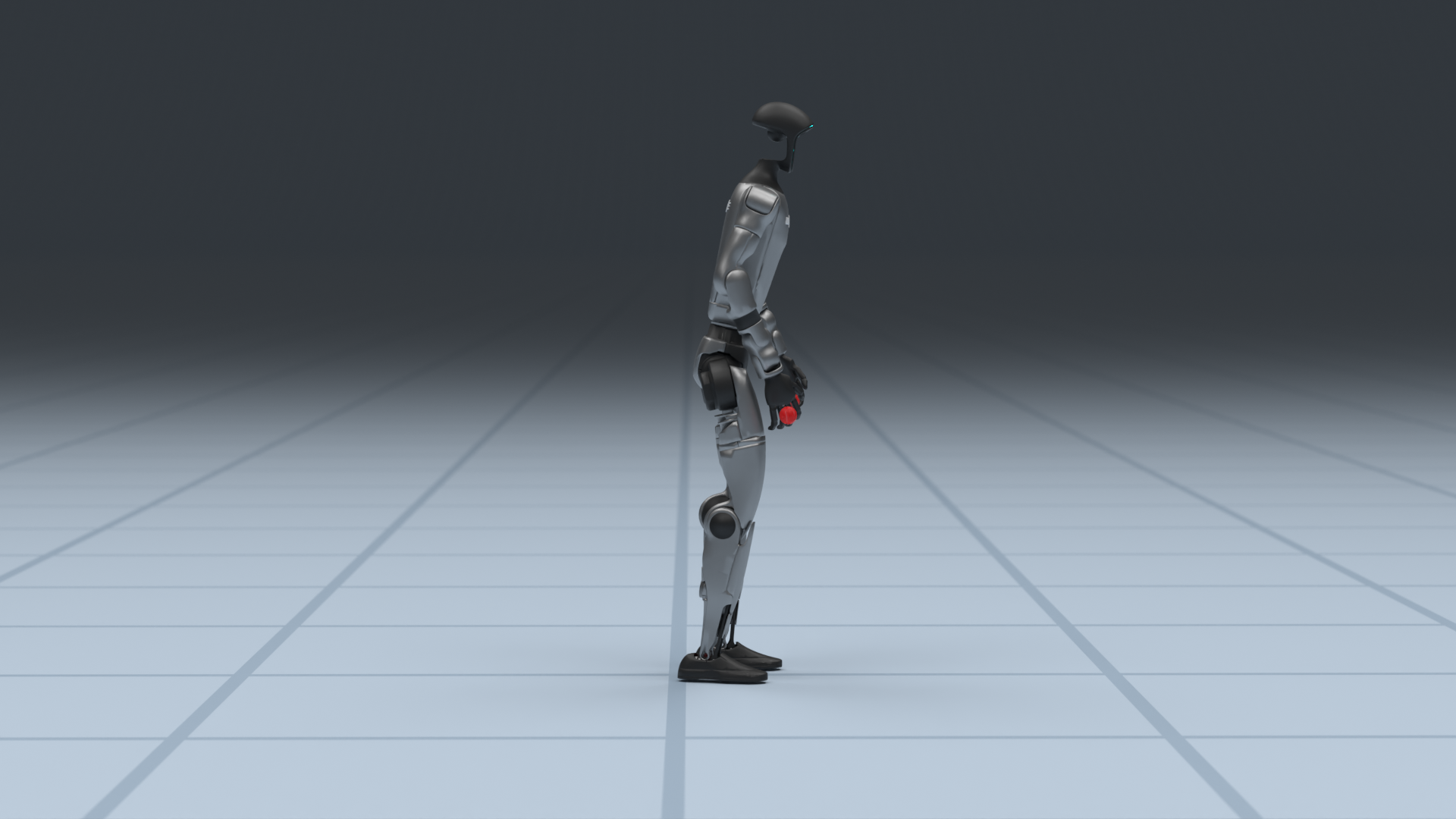}\hfill
       \includegraphics[trim={19cm 4cm 25cm 4cm},clip,width=0.248\textwidth]{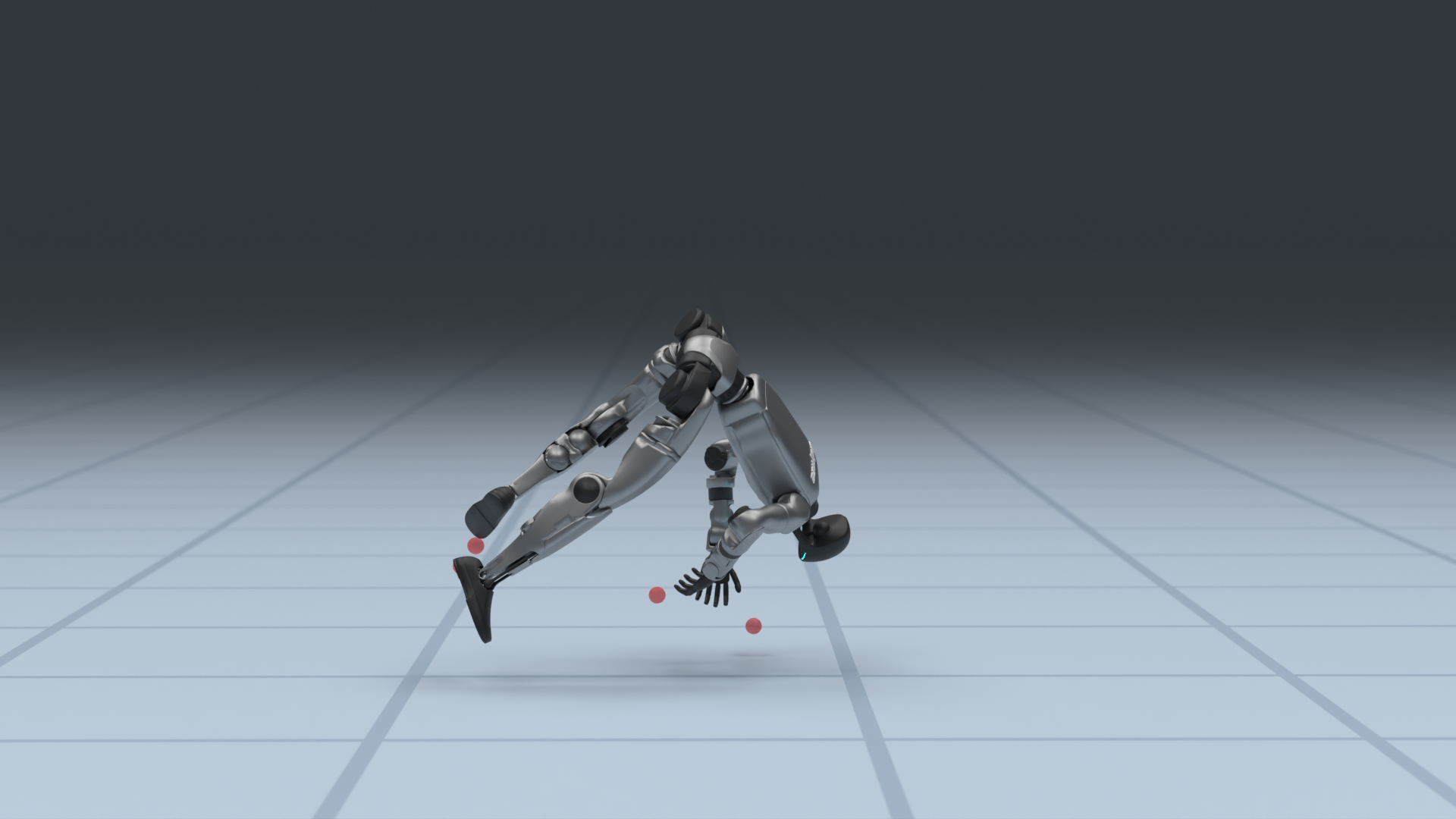}\hfill
       \includegraphics[trim={20cm 4cm 24cm 4cm},clip,width=0.248\textwidth]{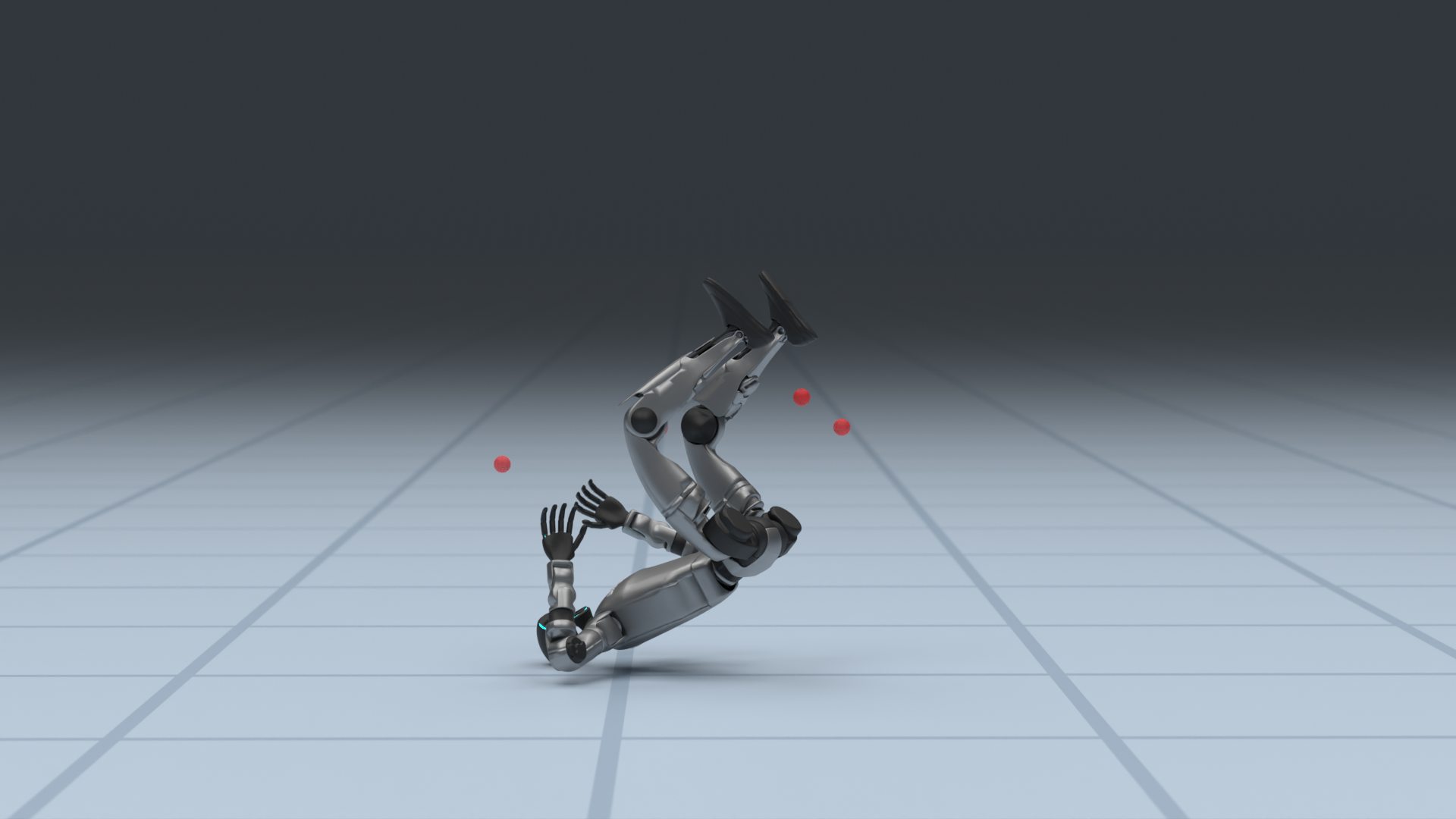}\hfill
       \includegraphics[trim={26cm 4cm 18cm 4cm},clip,width=0.248\textwidth]{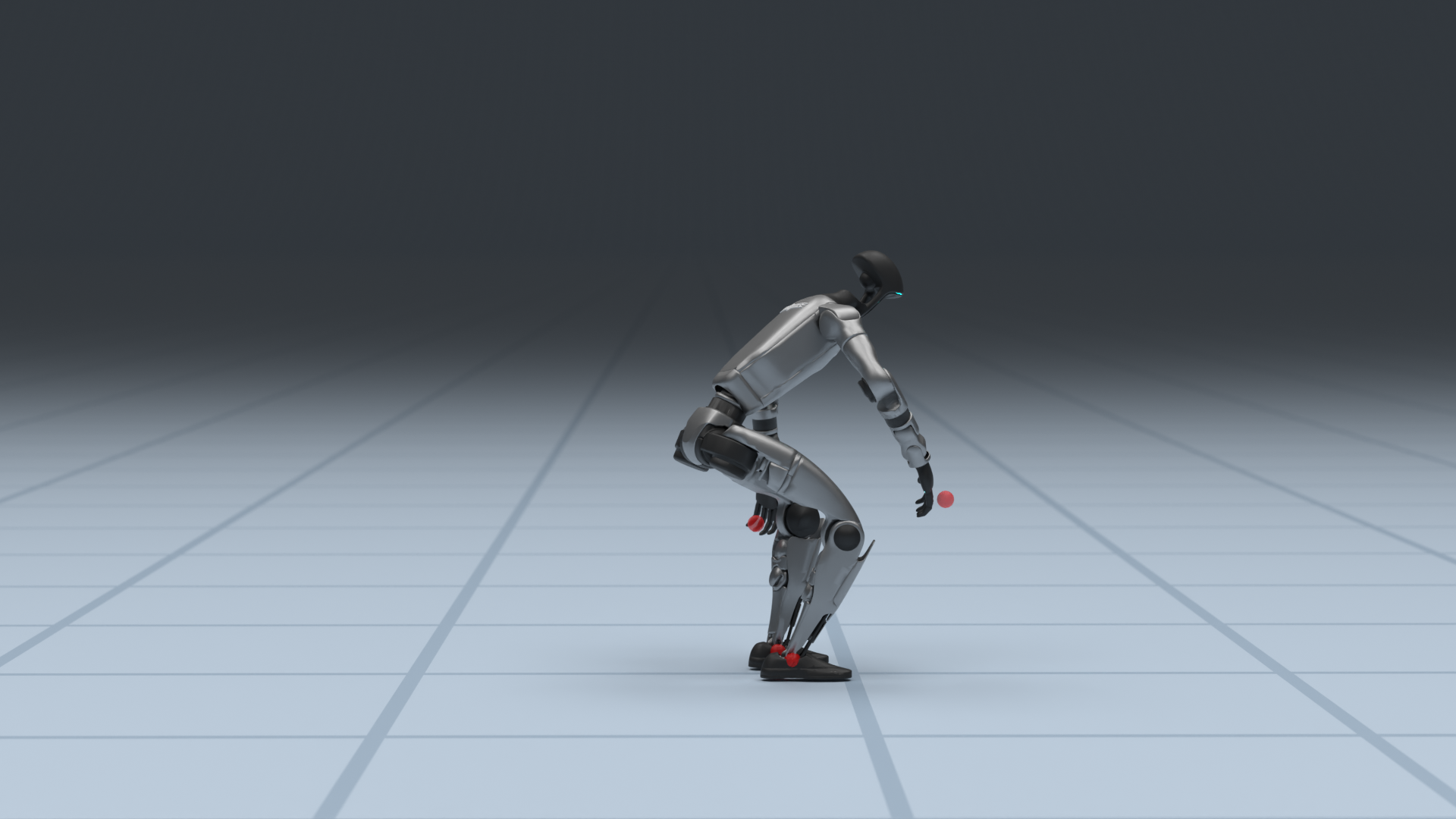}
       \textbf{Tracking (simulation):} Dense targets specify a front roll.
   \end{subfigure}
\caption{\textbf{Multiple control modes through one G1 policy.} Each mode specifies different goals for the same frozen controller.}
\label{fig: empirical}
\end{figure*}

\begin{figure}[!htbp]
    \centering
    \begin{subfigure}[b]{0.95\linewidth}
       \centering
       \includegraphics[trim={32cm 4cm 28cm 6cm},clip,width=\textwidth]{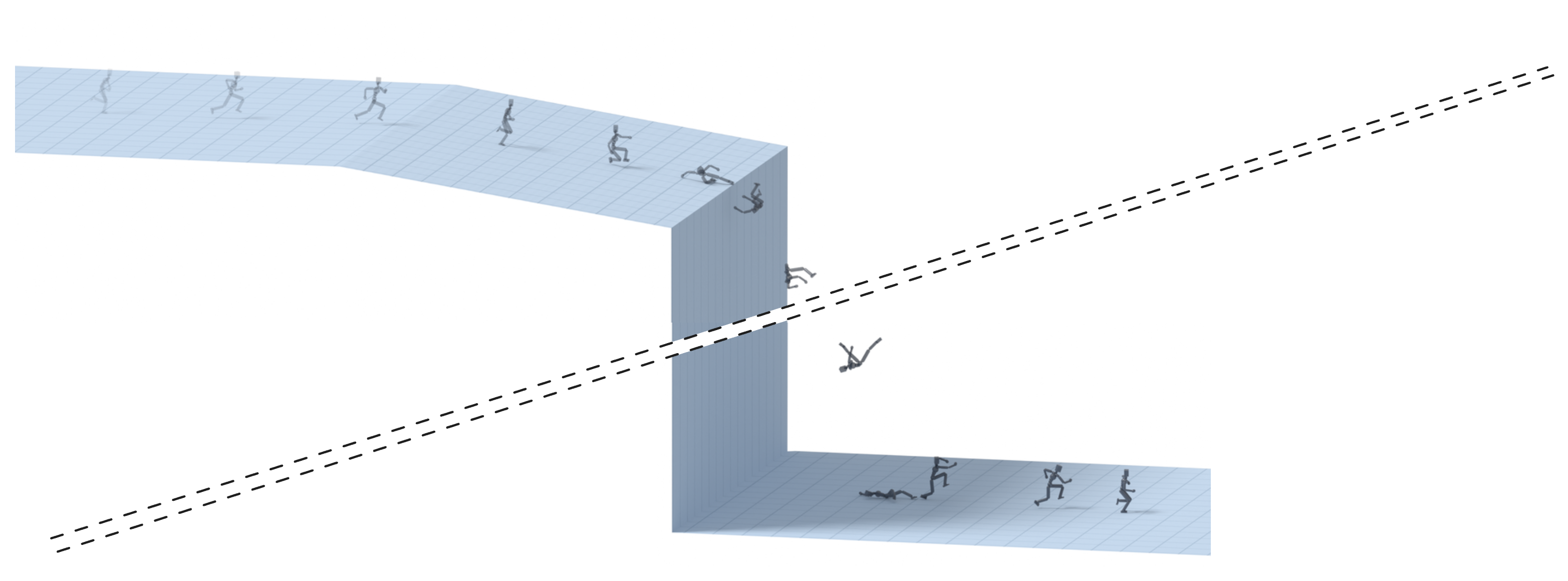}
   \end{subfigure}\\\vspace{0.2cm}
   \begin{subfigure}[b]{0.99\linewidth}
       \centering
       \includegraphics[trim={74cm 8cm 3cm 3cm},clip,width=\textwidth]{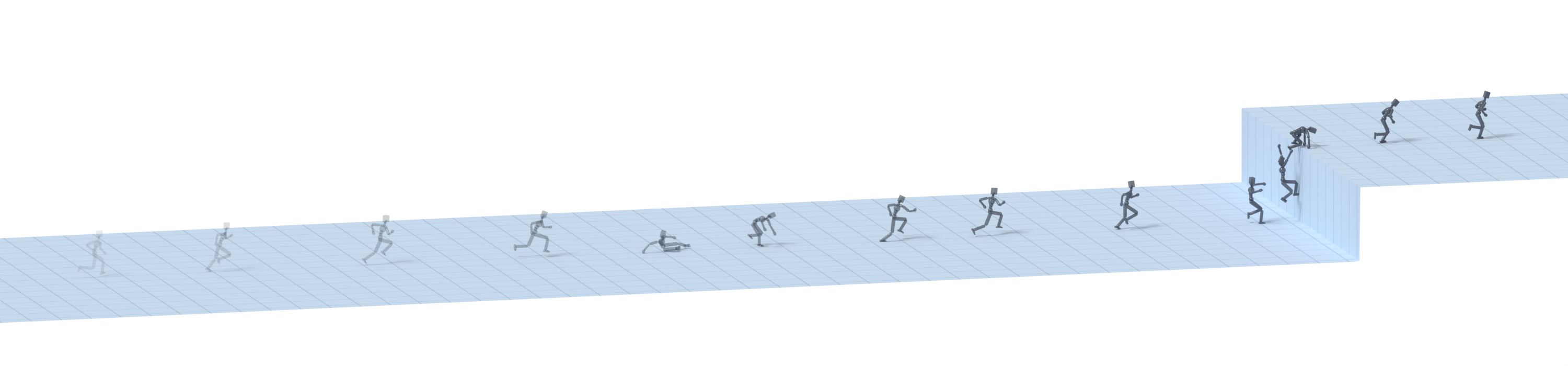}
   \end{subfigure}
\caption{\textbf{SOMA:} On a simulated SOMA character, with the same recipe, \method runs, slides off a ledge, recovers from the fall, then runs and climbs up a wall.}
\label{fig: soma}
\vspace{-0.2cm}
\end{figure}

\smallskip
\noindent\textbf{Deployment:}
Our C++ deployment stack compiles the base policy with TensorRT and CUDA graphs to run on a laptop with an NVIDIA GPU, which the robot connects to via Ethernet cable. An Xbox controller and a PICO headset provide mode selection, steering, and hand-pose commands. All hardware demonstrations use deterministic prior-mean inference. Automatic recovery uses FK-derived torso height and the state machine described in \Cref{sec:prompt-compiler}.

\subsection{Hardware Demonstrations}
\label{sec:qualitative}

\begin{figure*}[!t]
    \centering
   \begin{subfigure}[b]{0.49\linewidth}
       \centering
       \includegraphics[trim={0cm 3.5cm 0cm 3.5cm},clip,width=0.248\textwidth]{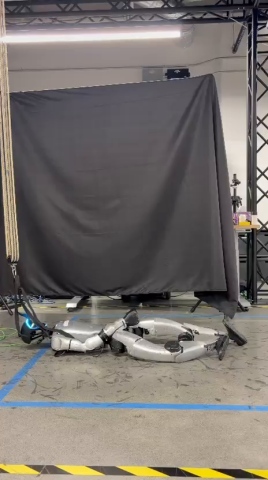}\hfill
       \includegraphics[trim={0cm 3.5cm 0cm 3.5cm},clip,width=0.248\textwidth]{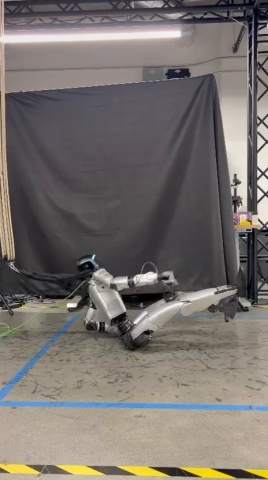}\hfill
       \includegraphics[trim={0cm 3.5cm 0cm 3.5cm},clip,width=0.248\textwidth]{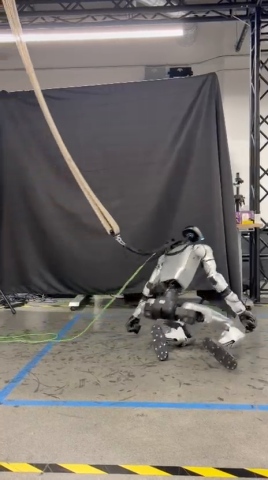}\hfill
       \includegraphics[trim={0cm 3.5cm 0cm 3.5cm},clip,width=0.248\textwidth]{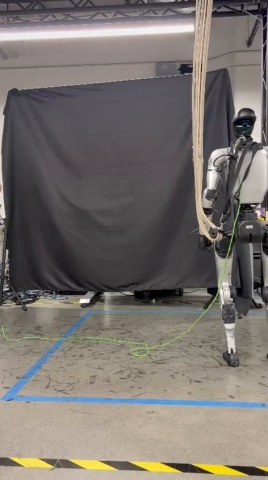}
       Lying on the back $\rightarrow$ sitting $\rightarrow$ standing
   \end{subfigure}\hfill
   \begin{subfigure}[b]{0.49\linewidth}
       \centering
       \includegraphics[trim={0cm 3cm 0cm 4cm},clip,width=0.248\textwidth]{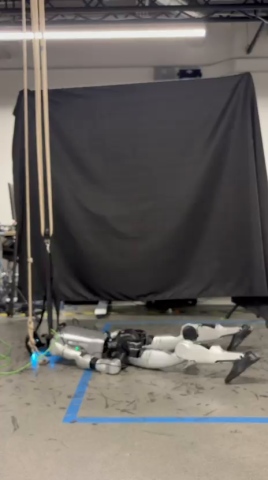}\hfill
       \includegraphics[trim={0cm 3cm 0cm 4cm},clip,width=0.248\textwidth]{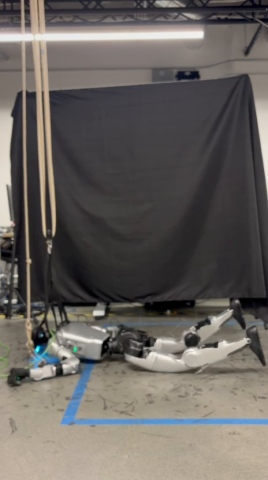}\hfill
       \includegraphics[trim={0cm 3cm 0cm 4cm},clip,width=0.248\textwidth]{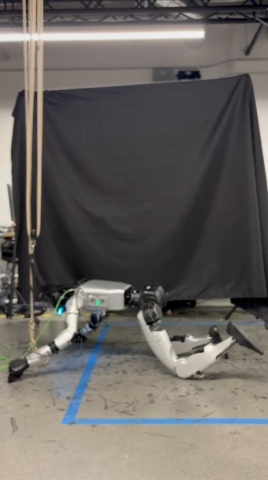}\hfill
       \includegraphics[trim={0cm 2cm 0cm 5cm},clip,width=0.248\textwidth]{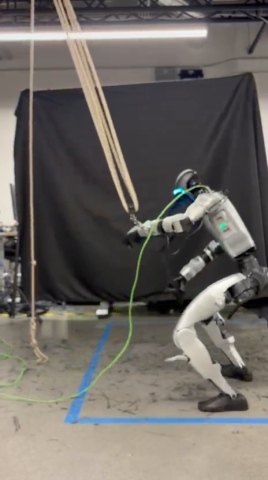}
       Lying on the front $\rightarrow$ hands and knees $\rightarrow$ standing
   \end{subfigure}\\\vspace{0.2cm}
   \begin{subfigure}[b]{0.49\linewidth}
       \centering
       \includegraphics[trim={0cm 3cm 0cm 5cm},clip,width=0.248\textwidth]{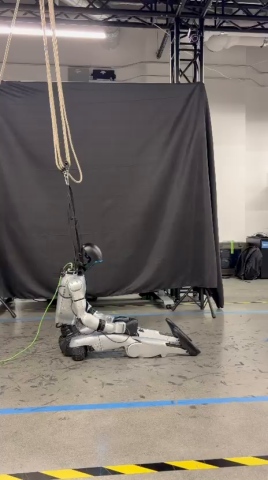}\hfill
       \includegraphics[trim={0cm 3cm 0cm 5cm},clip,width=0.248\textwidth]{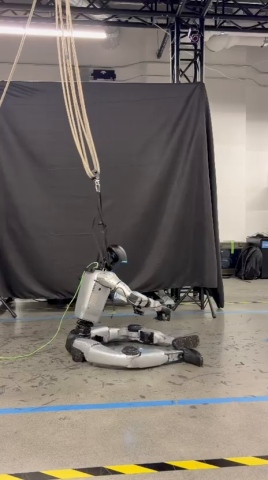}\hfill
       \includegraphics[trim={0cm 3cm 0cm 5cm},clip,width=0.248\textwidth]{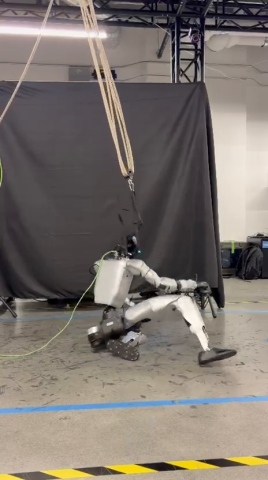}\hfill
       \includegraphics[trim={0cm 3cm 0cm 5cm},clip,width=0.248\textwidth]{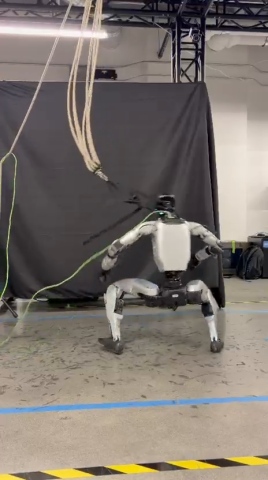}
   \end{subfigure}\hfill
   \begin{subfigure}[b]{0.49\linewidth}
       \centering
       \includegraphics[trim={0cm 2cm 0cm 5cm},clip,width=0.248\textwidth]{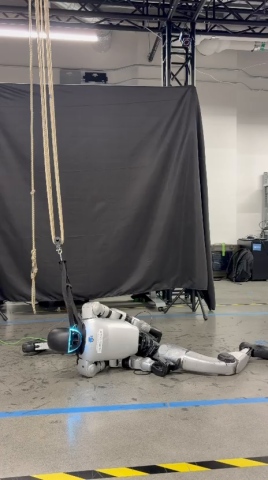}\hfill
       \includegraphics[trim={0cm 2cm 0cm 5cm},clip,width=0.248\textwidth]{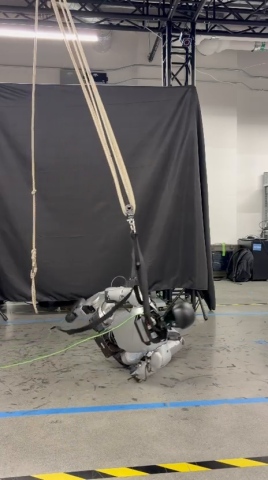}\hfill
       \includegraphics[trim={0cm 2cm 0cm 5cm},clip,width=0.248\textwidth]{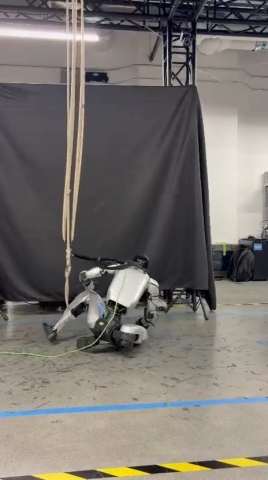}\hfill
       \includegraphics[trim={0cm 4cm 0cm 4cm},clip,width=0.248\textwidth]{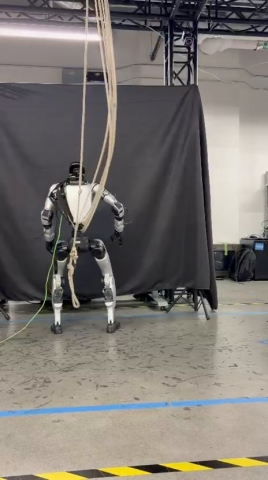}
   \end{subfigure}
\caption{\textbf{Recovery:} \method recovers from a range of initial configurations.}
\label{fig: recovery}
\end{figure*}

The physical G1 uses one frozen \method policy for all demonstrated control modes (\Cref{fig: empirical,fig: recovery}). Switching modes changes only the goal observations and masks; model weights and sensor processing remain unchanged.
%%JP: not sure if the part about "button press" is needed? Also, I'm not sure if we should be highlight the "one deployment stack" part here. The more imprtant thing to highlight seems to be "one model, multipel control modes"?

\smallskip
\noindent\textbf{Steering and gait refinement:}
We use an Xbox controller for interactive steering. One thumbstick controls heading, the other sets travel direction and speed. The steering skill adds timed foot-placement goals that produce more decisive steps and markedly improve walking quality, especially on hardware. Without these goals, the policy produces less committed foot swings and more hesitation. No policy retraining is required (\Cref{sec:prompt-compiler}).

\smallskip
\noindent\textbf{Teleoperation:}
A PICO headset provides steering via thumbsticks and maps the user's relative hand motions to target locations for the robot's hands. This model enables the user to direct the robot to pick objects, such as an apple.

\smallskip
\noindent\textbf{Sitting on objects:}
A skill can also be designed to enable the physical G1 to sit on a sofa and stand back up (\Cref{fig: empirical}). The robot is positioned near the sofa and the seat's bounding box is provided as an input condition. The skill then generates torso and foot targets.
%Autonomous furniture localization and approach are not evaluated.
%%JP: this part is not really necessary

% The physical G1 sits on a sofa and returns to standing (\Cref{fig: empirical}). We place the robot near the object and supply a cue that the object is nearby.
% %%JP: what is a cue in this context?
% This demonstrates sitting in a prepared spatial context,
% %%JP: not clear what "prepared spatial context" means
% not autonomous approach and sitting.

\subsection{Recovery and Compliance}

The generative controller can be prompted to exhibit human-like responsive behaviors. Temporally sparse goals leave flexibility in the motion between targets, allowing the robot to accommodate physical guidance during human--robot interactions. By training the controller with synthetic recovery data, the controller is able to get-up from diverse fallen configurations on the ground, including awkward configurations that are not recorded in the original mocap data. Hardware experiments show recovery from lying on the back, lying on the front, sitting, and other fallen poses (\Cref{fig: recovery}), extending the controller's repertoire beyond the original demonstrations. Recovery is triggered automatically when torso height falls below a safety threshold (\Cref{fig:recovery-fsm}).

\subsection{Simulation Capabilities}

To avoid potential damage to the physical G1 and its three-finger hands, we demonstrate highly dynamic skills in MuJoCo simulation. Dense body targets produce behaviors such as a front roll (\Cref{fig: empirical}). Sparse style frames bias the motion toward example poses: a torso $[x,y]$ target and arrival time specify where and when to reach, while five unordered crawling poses guide the style without imposing their original timing.

\smallskip
In addition to the simulated G1 model, we also train a policy for the SOMA character using a \method recipe similar to G1's. In \Cref{fig: soma}, a stateful skill supplies velocity, heading, and terrain-relative pelvis-height goals for locomotion. A low pelvis target and temporary right-foot target initiate sliding. After falling off the ledge, only velocity and heading remain through the fall and recovery. Climbing uses heading and a pelvis target above and beyond the ledge, leaving hand and foot placement unconstrained. The frozen policy executes the sequence without reference trajectories or full-body landing and recovery poses.

% Flush qualitative demonstrations before introducing quantitative comparisons.
\FloatBarrier
\subsection{Comparison with Reference-Tracking Pipelines}

A common alternative to a native whole-body controller is a reference-tracking pipeline: a learned motion generator produces a reference trajectory, and a dense tracker follows it. We compare \method against two such pipelines on the same G1 embodiment. The first uses ARDY~\cite{zhao2026autoregressive}, a state-of-the-art realtime autoregressive diffusion model that generates whole-body reference poses including hand keyframes. The second uses a fast GRU locomotion generator that we train to produce walking references; for reaching, we additionally overlay analytic arm IK on its output. Both pipelines use the same SONIC~\cite{luo2025sonic} dense tracker. For this comparison, \method is trained without synthetic recovery data (\cref{sec:training-data}).

\smallskip
\noindent\textbf{Whole-body reaching:} (\Cref{tab:reach}).
We evaluate wrist target reaching in Mujoco using 54 targets spanning three distances (0.35--1.8\,m, so the G1 will need to take steps), three heights (0.55--1.35\,m), and both hands; a target counts as reached if the wrist is held within 10\,cm while the robot remains upright. All methods that reach a target achieve comparable arm precision; what separates them is the reached rate. ARDY conditions on the hand target through classifier-free guidance rather than hard inpainting, so the held wrist oscillates across successive re-plans. The gap is largest at targets requiring whole-body adjustment: high reaches that demand counterbalance and low reaches that require a hip fold, where a generator that plans at a fixed standing height cannot adapt.

\smallskip
\noindent\textbf{Push robustness during steering:} (\Cref{tab:push}).
During a 1\,m/s forward walk, a force pulse is applied at the torso for 100\,ms, with directions spanning eight headings. Both reference-tracking pipelines have zero success above 600\,N, whereas \method succeeds in 75\% of trials at 800 and 1000\,N. This comparison measures the complete control pipelines; it does not isolate which training or interface choice causes the difference. The non-monotonic rates at 600--1000\,N should not be interpreted as improved robustness to stronger pushes.

\begin{table}[t]
    \centering
    \caption{Whole-body wrist reaching on a 54-target grid. \method is trained without synthetic recovery data for this comparison.
  A target is reached if the wrist is held within 10\,cm of it
  while the robot remains upright. Error is measured
  \textbf{only on reached targets}.}
    \label{tab:reach}
    \setlength{\tabcolsep}{4pt}
    \begin{tabular}{lccc}
        \toprule
        Method & Reached (\%) & Error (cm) & Upright (\%) \\
        \midrule
        \method & \textbf{83} & 5.9 & \textbf{100} \\
        ARDY + SONIC & 39 & 6.4 & 82 \\
        GRU + IK + SONIC & 28 & 5.6 & 81 \\
        GRU + SONIC & 0 & --- & 87 \\
        \bottomrule
    \end{tabular}
\end{table}

\begin{table}[t]
    \centering
    \caption{Push robustness during steering (100\,ms force pulse at torso, 8 directions). Success: upright after 8\,s. \method trained \textbf{without} recovery data. Results: success (\%).}
    \label{tab:push}
    \begin{tabular}{lccccc}
        \toprule
        & \multicolumn{5}{c}{Perturbation force (N)} \\
        \cmidrule(lr){2-6}
        Method & 200 & 400 & 600 & 800 & 1000 \\
        \midrule
        \method    & \textbf{100} & \textbf{100} & \textbf{62} & \textbf{75} & \textbf{75} \\
        ARDY+SONIC & \textbf{100} & \textbf{100} & 17 & 0 & 0 \\
        GRU+SONIC  & \textbf{100} & 75 & 25 & 0 & 0 \\
        \bottomrule
    \end{tabular}
\end{table}

\subsection{Ablation Study}

\begin{table}[t]
  \centering
  \caption{Push robustness during XY waypoint reaching. Horizontal torso pushes last 100\,ms, recurring every 1--2\,s during the first 5\,s until arrival. Success: reaching within 0.5\,m of the goal within 10\,s, with relative torso height $\geq 0.45$\,m and tilt $\leq 60^\circ$ at arrival. Entries report success (\%) with upper/lower 95\% Wilson CI deviations in percentage points over 500 episodes. Bold marks CI overlap with the highest rate at each force (unrounded bounds).}
  \label{tab:ablation}
  \small
  \setlength{\tabcolsep}{1.3pt}
  \begin{tabular}{@{}lccccc@{}}
      \toprule
      & \multicolumn{5}{c}{Perturbation force (N)} \\
      \cmidrule(lr){2-6}
      Method & 0 & 200 & 400 & 600 & 800 \\
      \midrule
      \method
      & $\boldsymbol{97.4^{+1.1}_{-1.8}}$
      & $\boldsymbol{94.8^{+1.6}_{-2.3}}$
      & $\boldsymbol{33.0^{+4.2}_{-4.0}}$
      & $\boldsymbol{7.4^{+2.6}_{-2.0}}$
      & $\boldsymbol{6.2^{+2.5}_{-1.8}}$ \\
      Global only
      & $87.6^{+2.6}_{-3.2}$
      & $87.8^{+2.6}_{-3.2}$
      & $\boldsymbol{38.2^{+4.3}_{-4.2}}$
      & $\boldsymbol{9.6^{+2.9}_{-2.3}}$
      & $\boldsymbol{6.2^{+2.5}_{-1.8}}$ \\
      Mocap only
      & $92.0^{+2.1}_{-2.7}$
      & $84.8^{+2.9}_{-3.4}$
      & $\boldsymbol{31.8^{+4.2}_{-3.9}}$
      & $\boldsymbol{6.4^{+2.5}_{-1.8}}$
      & $\boldsymbol{5.0^{+2.3}_{-1.6}}$ \\
      No aux.
      & $81.6^{+3.2}_{-3.6}$
      & $78.8^{+3.4}_{-3.8}$
      & $29.2^{+4.1}_{-3.8}$
      & $\boldsymbol{5.4^{+2.3}_{-1.7}}$
      & $\boldsymbol{4.6^{+2.2}_{-1.5}}$ \\
      MaskedMimic$^{+}$
      & $\boldsymbol{98.2^{+0.9}_{-1.6}}$
      & $90.0^{+2.3}_{-2.9}$
      & $27.8^{+4.1}_{-3.7}$
      & $4.2^{+2.1}_{-1.4}$
      & $\boldsymbol{4.6^{+2.2}_{-1.5}}$ \\
      HOVER$^{+}$
      & $66.8^{+4.0}_{-4.2}$
      & $71.0^{+3.8}_{-4.1}$
      & $\boldsymbol{30.0^{+4.2}_{-3.9}}$
      & $3.0^{+1.9}_{-1.2}$
      & $1.2^{+1.4}_{-0.6}$ \\
      \bottomrule
  \end{tabular}
  \vspace{-0.05cm}
\end{table}

We test which training components support robustness by requesting planar waypoint goals directly, without skill-generated foot targets or height constraints, under repeated horizontal torso pushes (\Cref{tab:ablation}). All models use the same training budget and matched parameter counts; the budget is given in the supplementary material. We evaluate 500 episodes per method and force, measuring upright arrival and reporting 95\% Wilson confidence intervals.

We compare \method with variants without recovery data (\emph{Mocap only}), without latent auxiliary losses (\emph{No aux.}), and with only global spatio-temporal prompts (\emph{Global only}). We also construct MaskedMimic$^{+}$~\cite{tessler2024maskedmimic} and HOVER$^{+}$~\cite{he2025hover} by adapting the original methods to our expert, training pipeline, motion-capture corpus, and compute budget, with substantially larger networks than their original implementations.

These adaptations serve as combined ablations of \method. MaskedMimic$^{+}$ combines all three ablations above and replaces our deterministic and sphere-projected stochastic streams with a single unbounded Gaussian latent. HOVER$^{+}$ further removes the conditional VAE and replaces the transformer with a deterministic MLP policy.

The full model achieves higher observed success than both the recovery-data and auxiliary-loss ablations at every tested force, with the largest absolute gains at 0--200\,N. It also exceeds \emph{Global only} at these forces, while \emph{Global only} has higher success at 400--600\,N and ties at 800\,N; their confidence intervals overlap at all three higher forces. Broader prompt support therefore remains competitive in this evaluation, although interval overlap does not establish equivalence. Compared with the adapted MaskedMimic$^{+}$ and HOVER$^{+}$ baselines, \method achieves higher observed success at every nonzero force, while MaskedMimic$^{+}$ leads slightly without pushes. However, several intervals overlap, and all methods fall below 10\% success at 600--800\,N, highlighting remaining limitations under strong perturbations.

% \begin{figure}[t]
%    \centering
%    \includegraphics[width=0.99\linewidth]{figures/ablation.png}
%    \caption{\textbf{Ablations under increasing perturbations.} Goal-reaching success rate versus the magnitude of horizontal velocity perturbations applied to the robot's anchor. Success requires reaching within $0.5\,$m of the target within $10\,$s. Each score is computed across $2,000$ episodes, with the error bars representing the $95\%$ Wilson confidence intervals \cite{wilson1927probable}.}
%    \label{fig:ablation}
% \end{figure}

% Keep experimental figures and tables with the results, before discussion.
\FloatBarrier
\section{Limitations}

\smallskip
\noindent\textbf{Motion coverage and conditioning:} \method's repertoire depends on training coverage, and underspecified goals or incidental style cues can produce unintended motions. For example, abruptly commanding $2\,\mathrm{m/s}$ from rest often triggers a front roll. Additional motion capture and targeted synthetic trajectories could broaden the repertoire, while refining skill-generated goals can improve motion selection without retraining. We smooth velocity goals with an exponential moving average to increase the requested speed gradually.

% \textbf{Motion coverage and conditioning.} \method's motion repertoire and supported commands depend on the training data and conditioning schemes. Training on additional motion-capture data and synthetic trajectories targeted at failure scenarios could broaden this repertoire, extending our approach to recovery-data generation. Underspecified goals and incidental speed or direction cues in style examples can still produce unintended motions. For example, abruptly commanding a velocity of $2\,\mathrm{m/s}$ while the robot is standing still often produces a front roll. We address this by smoothing the velocity goals with an exponential moving average (EMA), allowing the requested speed to increase gradually. Such refinements to skill-generated goals can guide the policy toward the intended behavior without additional training.

\smallskip
\noindent\textbf{Externally supplied goals:} The task system remains responsible for goal selection and long-horizon planning. We place the G1 near a chair or sofa and supply a nearby-object cue; autonomous approach followed by sitting is not demonstrated. Localization and perception could enable automatic goal generation, while LLM-generated programs could sequence goals for longer-horizon tasks~\cite{liang2023code}.

% \textbf{Dependence on externally supplied goals.} \method provides motion execution, while goal selection and long-horizon planning remain the responsibility of the task system. Our sitting demonstrations illustrate this distinction: we position G1 near a chair or sofa and supply a nearby-object cue, rather than demonstrating autonomous approach followed by sitting. Integrating localization and object perception would provide the spatial information needed to generate these goals automatically. Progress in LLM-guided execution suggests a complementary direction: generating and sequencing goals through the prompt interface to coordinate longer-horizon tasks while keeping the motion controller frozen \cite{brohan2023can,liang2023code}.

\smallskip
\noindent\textbf{Hardware evaluation:} Our hardware study demonstrates diverse behaviors with a single frozen controller, including recovery from tested configurations and qualitative gait improvements. Trials across initial and operating conditions would quantify recovery reliability and gait stability and assess contact safety, actuation limits, and failure modes.

% \textbf{Limited hardware evaluation.} Hardware get-ups demonstrate recovery for the tested configurations, not reliable recovery from arbitrary falls. The observed gait improvements are qualitative; we have not quantified their effect on hardware stability. Broader evaluation is needed to measure recovery success across initial conditions and characterize contact safety, operation near torque limits, and failure modes.
\section{Conclusion}

\method converts dense motion supervision into a generative whole-body controller for partial goals. Synthetic recovery data expand its repertoire, and body-level prompts support runtime control in simulation and on G1. Within the learned repertoire, new tasks require goal-generating programs, not policy retraining. Skills specify goals, timing, and transitions; the frozen policy coordinates the body.

% No outstanding float should interrupt or follow the bibliography.
\FloatBarrier
\bibliographystyle{IEEEtran}
\bibliography{references}

\ifincludeappendix
\clearpage
\appendices
\section{Implementation Details}
\label{sec:appendix}

This appendix separates three stages: PPO trains motion-tracking experts; supervised distillation trains the prompt-conditioned student; runtime skills construct prompts for the frozen student. We give their settings below, followed by observation construction and deployment details. Notation follows \cref{sec:method}.

\subsection{Training Setup}
\label{sec:app-training}

We train in ProtoMotions~\cite{tessler2025protomotions} using Isaac Lab~\cite{mittal2025isaaclab}. G1 distills four frozen experts into one student: BONES~\cite{bonesstudio}, AMASS+LAFAN, recovery, and SAMP. These groups use 12, 4, 2, and 2 GPUs, respectively. Each GPU simulates 2048 environments, collects 32 control steps per rollout, and optimizes minibatches of 4096 samples for two passes. Averaging equal-sized batches across GPUs gives the groups relative training weights $(0.6,0.2,0.1,0.1)$. All GPUs share the student parameters; each group uses its own expert, motions, and scenes.

G1 simulates at 200 Hz with four physics steps per action (50 Hz control); SOMA uses 120 Hz physics and 30 Hz control. Runtime skills use the trained student without further weight updates.

\paragraph{Compute budgets}
Student training uses 20 GPUs. The ablations use eight A100 GPUs for $5{,}000$ training iterations with matched parameter counts. These budgets exclude expert training.

\paragraph{Training data}
The dataset includes BONES motion capture and synthetic recovery motions; G1 additionally uses AMASS+LAFAN and SAMP. The SOMA scene dataset contains wall-climbing and ground-sliding motions with collision geometry.

\subsection{Dense Reference Trackers}
\label{sec:app-tracker}

\paragraph{Data and observations}
The SOMA teacher learns from flat-ground motion capture, recovery, climbing, and sliding. Walls and ramps are included in the collision geometry, and a local $16\times16$ height map provides terrain context. G1's BONES, AMASS+LAFAN, and recovery groups use flat ground; SAMP supplies scene interactions. Trackers observe privileged body positions, orientations, and velocities, together with complete reference poses sampled at future times.

\paragraph{G1 BONES expert}
The G1 BONES expert uses finite scalar quantization (FSQ). An encoder with hidden widths $(2048,1024,512,512)$ maps future poses of the torso, hands, and feet, together with joint targets, to 64 latent coordinates with 33 quantization levels each. Its actor decoder predicts the mean of a Gaussian action distribution; a separate critic predicts state value. Both have two hidden layers at each width, in order: 4096, 2048, 1024, and 512, with SiLU activations. These eight-layer networks belong to the expert; the deployed student's decoder has three 1024-unit layers. The expert outputs 29 joint commands.

The G1 BONES expert learns one log-standard-deviation vector per motion, shared across states of that motion. The bounds are $\log(0.001)$ and $\log(0.5)$, its learning rate is $5\times10^{-4}$, and the entropy coefficient is $0.01$. This table controls PPO exploration; the distilled student does not inherit it.

\paragraph{SOMA expert}
The SOMA teacher has eight 1024-unit actor hidden layers, six 1024-unit critic hidden layers, and 66 action outputs. It uses fixed log-standard deviation $-2.9$. Its smoothness penalty acts on consecutive joint-position commands: $-0.1\|q_t^{\mathrm{des}}-q_{t-1}^{\mathrm{des}}\|_2$.

\paragraph{Expert optimization}
Both experts use PPO with discount $0.99$, generalized advantage estimation (GAE) parameter $0.95$, and ratio clipping $0.2$. Actor and critic Adam learning rates are $2\times10^{-5}$ and $10^{-4}$, with momentum coefficients $(0.95,0.99)$. Both collect 32-step rollouts. \Cref{tab:expert-ppo} lists the differing settings. G1 also uses clipped value loss.

\begin{table}[!htbp]
\centering\small
\caption{Expert PPO settings, distinct from student distillation.}
\label{tab:expert-ppo}
\begin{tabular}{@{}lrr@{}}
\toprule
Setting & G1 BONES & SOMA \\
\midrule
Environments per GPU & 8192 & 4096 \\
Minibatch size per GPU & 16384 & 8192 \\
Optimization passes per rollout & 2 & 1 \\
Gradient-norm limit & 1 & 50 \\
Entropy coefficient & 0.01 & 0.005 \\
Reward normalization & No & Yes \\
\bottomrule
\end{tabular}
\end{table}
G1 rewards are multiplied by $0.8$ before PPO optimization.

\begin{table}[t]
\centering\small
\caption{G1 BONES expert rewards. Tracking terms are $w\exp(-e/\sigma^2)$. Position and velocity errors sum squared coordinate errors, then average over the selected bodies; rotation error is the mean squared relative rotation angle.}
\label{tab:g1-tracker-rewards}
\begin{tabular}{@{}lrr@{}}
\toprule
Term & $w$ & $\sigma$ \\
\midrule
Torso, hands, feet: global position & 1 & 0.3 \\
Torso, hands, feet: global rotation & 1 & 0.4 \\
Torso and hands: global position & 2 & 0.1 \\
Torso and hands: global rotation & 0.4 & 0.1 \\
Anchor-relative body position & 1 & 0.3 \\
Anchor-relative body rotation & 1 & 0.4 \\
Body linear velocity & 1 & 1.0 \\
Body angular velocity & 1 & 3.14 \\
\midrule
Soft joint-limit violation & $-10$ & --- \\
Raw action squared difference & $-0.1$ & --- \\
Contact mismatch & $-0.1$ & --- \\
Self-contact count & $-0.1$ & --- \\
\bottomrule
\end{tabular}
\end{table}

\paragraph{Speed-dependent precision}
The SOMA teacher uses speed-gated precision rewards; the G1 BONES expert in \cref{tab:g1-tracker-rewards} does not. For SOMA, let $\bar v_b$ be the average over selected future samples of reference displacement magnitude divided by time offset, and let $\bar\omega_b$ use relative rotation angle instead of displacement. The per-body gates are
\begin{equation}
 \eta_b^{p}=\exp[-(\bar v_b/\rho_p)^2],\qquad
 \eta_b^{R}=\exp[-(\bar\omega_b/\rho_R)^2],
 \label{eq:slow-gate}
\end{equation}
with $\rho_p=0.5\,$m/s and $\rho_R=1\,$rad/s. For a selected body set $B$, the reward is
\begin{equation}
 r_B^k=\exp\left[-\frac{\lambda_k}{|B|}
                    \sum_{b\in B}\eta_b^k e_b^k\right].
\end{equation}
Position error is $e_b^p=\|p_b-p_b^*\|^2/3$; rotation error is the squared relative rotation angle. The mean is taken \emph{inside} the exponential. Position and rotation sensitivities are 5000 and 200. Hand terms each have weight 1; foot, anchor, and head terms each have weight $0.5$. At high reference speeds the gates approach zero, flattening the reward rather than increasing its weight. The teacher also uses ordinary position, rotation, linear-velocity, angular-velocity, and root-height terms with $(w,\lambda)=(0.5,100),(0.3,5),(0.1,0.5),(0.1,0.1),(0.2,100)$, processed-command smoothness, and a power penalty. These tracker rewards are distinct from the student's supervised losses.

\subsection{Masked On-Policy Distillation}
\label{sec:app-distillation}

\paragraph{Prior encoder}
Separate encoders map the prompt streams and sensor history into 1024-dimensional tokens. The transformer has six layers, four attention heads, 4096-dimensional feed-forward blocks, and zero dropout. Separate $(512,256,128)$ MLP heads predict the 64-dimensional condition $d_t$, prior mean, and prior log-variance. Masks are explicit inputs; unavailable tokens are excluded by attention and invisible style frames by masked pooling. A zero-valued visible target remains different from an absent target.

\paragraph{Fusion and action decoder}
The fusion MLP maps $[d_t,z_t]$ to the 64-dimensional representation $h_t$ using hidden widths $(512,256,128)$. The action decoder combines $h_t$ with normalized current observations and the previous action, then applies three 1024-unit hidden layers. G1 uses body poses reconstructed by forward kinematics (FK), joint measurements, and root-state measurements, and outputs 29 commands. SOMA uses simulator-provided body states and a height map, and outputs 66 commands. G1 uses SiLU and SOMA uses ReLU.

\paragraph{Training-only networks}
The privileged posterior uses five 1024-unit hidden layers and a 512-dimensional output, followed by separate mean-correction and log-variance heads with hidden widths $(512,256,128)$. It receives full reference features and prompt features. The dynamics predictor maps $h_t^q$ to the next condition $d_{t+1}$ through hidden widths $(512,256,128)$. SOMA supplies height-map and root-height information to both the prior and posterior.

The prompt encoders use three 1024-unit hidden layers; style frames share an encoder and are pooled over the visible frames. The posterior's canonicalized- and world-target encoders use $(512,512)$ hidden layers with 256-dimensional outputs. G1 additionally encodes scene-box corners with a 1024-unit hidden layer and 1024-dimensional output. Normalized observations are clipped to $[-5,5]$; masks are not normalized. Inference uses the normalization statistics learned during training.

\paragraph{Sampling and rollouts}
At reset, each environment draws one Gaussian noise vector and retains it for the episode. The prior and posterior share this vector, but their means and variances change with the observations and prompt. Their latent samples therefore remain context-dependent. Each sample is projected onto the unit sphere before fusion with $d_t$.

At the same reset, control is assigned to the posterior-guided student with probability $0.9$ and to the frozen tracker otherwise. That controller acts for the whole episode. The tracker labels every visited state under either controller, providing supervision on both student- and expert-induced states. At deployment, actions use the prior path.

\paragraph{Losses}
In \cref{eq:distillation-loss}, $\lambda_{\mathrm{dyn}}=\lambda_{\mathrm{tc}}=\lambda_\mu=10^{-3}$ and $\lambda_{\mathrm{act}}=5\times10^{-2}$. Both students use Adam with learning rate $2\times10^{-5}$, momentum coefficients $(0.9,0.999)$, $\epsilon=10^{-8}$, no weight decay, and gradient-norm clipping at 50. The imitation loss averages squared differences between student and expert actions over action coordinates, before conversion to joint-position commands. Student training uses supervised losses rather than PPO rewards.

\paragraph{Gaussian KL and spherical decoding}
Write $\ell^p_j=\log\sigma_j^2$ and $\ell^q_j=\log{\sigma_j'}^2$. The implemented per-sample KL is
\begin{equation}
 D_{\mathrm{KL}}(q\|p)=\frac12\sum_{j=1}^{64}
 \left[\ell^p_j-\ell^q_j+
 \frac{\exp(\ell^q_j)+(\Delta\mu_j)^2}{\exp(\ell^p_j)}-1\right].
 \label{eq:app-kl}
\end{equation}
The KL is summed over latent coordinates and averaged over samples. The posterior predicts its own log-variance and a residual mean relative to the prior. Sampling uses $\tilde z^q=\mu+\Delta\mu+\exp(\ell^q/2)\odot\epsilon$ for the posterior and $\tilde z^p=\mu+\exp(\ell^p/2)\odot\epsilon$ for the prior. KL and mean-norm regularization act on these Gaussian parameters; the sampled latent is normalized to unit length before decoding.

\paragraph{KL schedule}
G1 uses cyclical annealing. With epoch $e$, define
\begin{equation}
 \begin{aligned}
 \hat e&=\begin{cases}e,&e\leq2000,\\
 500+((e-500)\bmod1501),&e>2000,\end{cases}\\
 \beta_e&=10^{-4}+(10^{-2}-10^{-4})
       \operatorname{clip}\left(\frac{\hat e-500}{1500},0,1\right).
 \end{aligned}
 \label{eq:app-cyclical-kl}
\end{equation}
The weight rises from $10^{-4}$ to $10^{-2}$ over epochs 500--2000, then repeats the ramp. SOMA uses the same initial ramp and holds $10^{-2}$ thereafter. An epoch denotes one rollout followed by optimization.

\paragraph{Temporal targets}
Temporal losses use consecutive steps within the same episode and exclude pairs that cross a reset. The dynamics loss predicts the next condition $d_{t+1}$ from $h_t^q$; consistency and action smoothing compare the current representation or posterior action with its previous value. Targets are stored during rollout and treated as constants during optimization (stop-gradient). Pairs are formed before minibatch shuffling, and errors are averaged over coordinates and valid pairs. Action smoothing compares student predictions even in expert-controlled episodes. These losses use per-step differences without division by $\Delta t$.

\paragraph{SOMA latent diversity}
SOMA additionally encourages different latents to produce different actions; G1 omits this loss. Holding the observation and condition fixed, draw two independent unit-sphere latents $z_1,z_2$ and decode actions $a_1,a_2$. With $c(a)=\operatorname{clip}(a,-1,1)$ and $A$ action coordinates,
\begin{equation}
 \mathcal L_{\mathrm{div}}=10^{-4}\,\mathbb E\left[
 \left(1-\frac{A^{-1}\|c(a_1)-c(a_2)\|_2^2}
             {\tfrac12(1-z_1^\top z_2)+10^{-5}}\right)^2\right].
\end{equation}
The square is applied per sample before averaging. Both decoded actions receive gradients; the sampled latents and conditioning inputs do not. This term supplements, rather than replaces, the temporal losses.

\paragraph{Training iteration}
At reset, choose the motion and prompt masks, sample latent noise for the episode, and choose student or expert control. At each step, compute both the expert label and the student's posterior action, execute the chosen controller, and record the transition. After a 32-step rollout, form within-episode temporal pairs, shuffle transitions into minibatches, and perform two passes of supervised updates. Student gradients are averaged across GPUs; experts remain frozen. Inference uses only the prior, fusion network, and action decoder.

\subsection{Prompt and Style-Cue Sampling}
\label{sec:app-sampling}

Training varies both the information in a prompt and its time horizon. For each body-target stream and episode, a spacing $\Delta\in[0.1,2.0]\,$s places five future slots at $t+k\Delta$, $k\in\{1,\ldots,5\}$. Dense, intermediate, and sparse schedules are chosen with probabilities $(0.2,0.6,0.2)$, each using its own spacing subrange. Dense schedules expose every slot, sparse schedules expose one, and intermediate schedules expose each slot with probability $0.8$. Masks also select body identities and target properties. Position targets support separate planar $(x,y)$ and vertical $z$ constraints. Anchor orientation can be specified as heading alone or as a full three-dimensional rotation; position and orientation goals can be selected independently.

The world-target, canonicalized-target, and planar-velocity streams are each active with probability $0.5$. The style cue samples up to five frames from a two-second sliding window. Its visible-frame-count distribution, for zero through five frames, is $(0.50,0.15,0.15,0.10,0.07,0.03)$, so it too is absent in half the episodes. The four presence decisions are independent, making all sixteen combinations equally likely, including no motion goal.

Scene context is separate from these four masks. G1 accepts one object represented by up to two boxes. Valid boxes remain visible in the SAMP group; unused boxes and all box tokens in scene-free groups are masked out. This fixed input shape lets the same student operate with or without furniture. SOMA instead receives scene geometry through its local height map.

Each style frame is canonicalized like a local target pose. Heading-aligned linear and angular body velocities are included with probability $0.5$. The frames carry neither arrival times nor global placement. A shared frame encoder followed by masked mean pooling makes the style representation invariant to their order. This distinguishes an example-based distribution cue from timed body targets.

\subsection{Observation Construction}
\label{sec:app-observations}
\label{sec:app-g1-world}

\paragraph{SOMA}
The simulator provides body positions, orientations, and velocities, root height, and a local $16\times16$ height map. These inputs support both current-state observations and comparisons with world-frame goals.

\paragraph{G1 current pose}
Let $A$ denote the torso anchor, $R_t^A$ its orientation reconstructed from IMU and joint measurements, and $H_t$ the yaw-only rotation extracted from $R_t^A$. The G1 sensor suite places the IMU on the pelvis; torso orientation is derived through FK rather than treated as an independent torso-mounted measurement. FK applied to measured joint angles $q_t$ gives each body's torso-relative position $p_b(q_t)$ and orientation $R_b(q_t)$. Rotating into the heading frame gives $\bar p_t^b=H_t^\top R_t^A p_b(q_t)$. We then set the lowest body point to zero height:
\begin{equation}
 \begin{aligned}
 \gamma_t&=\min_{j\in\mathcal J}\bar p_{t,z}^j,\\
 \hat p_t^b&=\bar p_t^b-\gamma_t e_z,\\
 \hat R_t^b&=H_t^\top R_t^A R_b(q_t),\qquad b\in\mathcal B,
 \end{aligned}
 \label{eq:g1-current-fk}
\end{equation}
where $\mathcal J$ contains all bodies, $\mathcal B$ contains the torso, hands, and feet used for prompts, and $e_z$ is the vertical unit vector. With a 6D rotation representation, this gives nine pose features per prompted body. This normalization defines a local pose convention; it does not estimate ground height.

The history encoder receives five steps of joint position, joint velocity, root orientation, angular velocity in the root frame, and reconstructed torso orientation, together with the time elapsed since each measurement. The current-state encoder and action decoder also receive the previous action. Sensor noise is applied before FK: independent uniform perturbations have half-widths $0.01\,$rad for joint position, $0.5\,$rad/s for joint velocity, $0.05$ per IMU quaternion component, and $0.2\,$rad/s for gyroscope readings. The perturbed quaternion is renormalized. Localization, base velocity, and ground clearance receive no added noise during training.

\paragraph{Physical randomization}
All four G1 groups add action offsets in $[-0.025,0.025]$ and use 64 sampled material settings with static friction in $[0.3,1.6]$, dynamic friction in $[0.3,1.2]$, and restitution in $[0,0.5]$. Torso center-of-mass offsets span $\pm0.025\,$m in $x$ and $\pm0.05\,$m in $y,z$. Hand masses are sampled log-uniformly between $0.001$ and $1.0\,$kg. Pushes are implemented as velocity perturbations every $1$--$3\,$s, with per-axis linear bounds $(0.5,0.5,0.2)\,$m/s and angular bounds $(0.52,0.52,0.78)\,$rad/s. These training perturbations differ from the force pulses used in evaluation. G1 training adds no sensor or action latency. SOMA disables domain randomization.

\paragraph{Canonicalized targets}
Each target pose uses its own torso position and heading at target time $\tau$. Let $(p_\tau^{*,b},R_\tau^{*,b})$ be the reference body poses and $H_\tau^*$ the target torso heading. With $\gamma_\tau=\min_{j\in\mathcal J}p_{\tau,z}^{*,j}$, the target features are
\begin{equation}
 \begin{aligned}
 p_{\tau,xy}^{\ell,b}
   &=\bigl[H_\tau^{*\top}(p_\tau^{*,b}-p_\tau^{*,A})\bigr]_{xy},\\
 p_{\tau,z}^{\ell,b}&=p_{\tau,z}^{*,b}-\gamma_\tau,\\
 R_\tau^{\ell,b}&=H_\tau^{*\top}R_\tau^{*,b}.
 \end{aligned}
 \label{eq:g1-local-target}
\end{equation}
These targets and the current FK view are encoded separately for the network to compare. Training canonicalizes the complete target pose before masking bodies and properties, so the vertical origin remains well defined. A runtime skill uses the same convention when constructing a target. Canonicalized target values depend on the requested pose, not the current robot state.

\paragraph{World-frame targets}
A localization estimate $\tilde p_t^A$ places the FK pose in the world: $\tilde p_t^b=\tilde p_t^A+R_t^A p_b(q_t)$ and $\tilde R_t^b=R_t^A R_b(q_t)$. Target features then encode position differences in the current heading frame and relative rotations:
\begin{equation}
 \begin{aligned}
 \delta p_{t,\tau}^b&=H_t^\top(p_\tau^{*,b}-\tilde p_t^b),\\
 \delta R_{t,\tau}^b&=(\tilde R_t^b)^\top R_\tau^{*,b}.
 \end{aligned}
 \label{eq:g1-global-deltas}
\end{equation}
Rotations use the 6D representation, and masks select the requested properties. World-position deltas require the target and robot poses to share an origin: planar targets need $x$--$y$ localization, and absolute-height targets also need vertical registration. Subtracting the same floor offset from both poses leaves their difference unchanged. Canonicalized targets omit global placement and do not require this registration.

\subsection{Recovery-Data Generation}
\label{sec:app-recovery}

We extend HIL~\cite{wang2026hil} with MaskedMimic-style spatio-temporal conditioning~\cite{tessler2024maskedmimic}, masking body constraints and future times. The trained policy generates five-second recovery clips offline from randomized ground poses, using $1024$ parallel environments.

\paragraph{Initial states}
We sample a root orientation and joint targets, simulate 150 physics steps, and zero all velocities. The sampled poses thus settle under contact dynamics before recording begins.

\paragraph{Targets and acceptance}
Each rollout receives a sparse timed target from a source motion, with at least one position-constrained body. During collection, termination conditions are disabled. A clip is accepted if it contains no reset and its final mean Euclidean error over position-constrained bodies is below $0.25\,$m. This filter checks endpoint tracking, not sustained balance, contact quality, or hardware safety.

\paragraph{Storage and use}
Accepted clips store body and joint poses, velocities, and contacts. We subtract only the initial root $x$--$y$ translation, preserving height and yaw. The clips enter the motion library in the same format as captured motions and supervise the same tracking and distillation stages. The generator is used only for data collection; it is not part of the deployed controller.

\subsection{Harness Skills}
\label{sec:app-skills}

Each skill is a stateful function or finite-state machine that constructs prompts from task input. The harness runs these skills and manages transitions between them. \Cref{tab:app-skills} summarizes the implemented skills; none requires training a new motor policy.

\paragraph{Steering trajectories}
For the rule in \cref{sec:prompt-compiler}, the fitted coefficients are $A(v)=0.75(0.13+0.065v)$, $f(v)=\min(2,0.83+0.93v)$, $W=0.12$\,m, and $H=0.08$\,m, with speed in m/s and cadence in Hz. Here $W$ controls lateral widening, not the nominal stance width. At each control step, phase advances by $2\pi f(v)\Delta t$. The skill evaluates foot targets at arrival times $\tau\in\{0.1,0.2,0.3,0.4,0.5\}\,$s and transforms them from the planned torso frame into world coordinates using the torso position and heading targets at each arrival time.

Deployment additionally enforces a nominal stance half-width of at least $0.07\,$m. At or below $0.05\,$m/s, stationary-support targets replace the periodic gait. The skill specifies body-position goals; the learned policy determines joint commands and physical contacts.

\begin{table*}[!t]
\centering
\caption{Harness skills and the prompts they construct. Each embodiment uses
one frozen policy across its supported skills. The recovery row describes the
simulation state machine; hardware get-up demonstrations are reported in
\cref{sec:qualitative}.}
\label{tab:app-skills}
\small
\setlength{\tabcolsep}{3pt}
\begin{tabular}{@{}p{0.13\textwidth}p{0.20\textwidth}p{0.40\textwidth}p{0.17\textwidth}@{}}
\toprule
Skill & Input & Prompt fields & Update \\
\midrule
Dense tracking & reference clip & timed body positions and orientations & reference clock \\
Steering & joystick velocity and turn & planar velocity; five timed torso, heading, and alternating foot targets & every step \\
Path following & waypoint list & timed torso waypoints and planar velocity & receding horizon \\
Location & target position or orientation & timed body constraint; planar position, height, heading, or full anchor orientation & on target change \\
Style & example clip & unordered motion-state cue & on reset or override \\
Sparse goal & pose from a trajectory & one timed future pose & on target transition \\
Recovery (simulation) & torso height & canonicalized upright pose for all bodies; velocity and style masked & height-triggered transitions \\
\bottomrule
\end{tabular}
\end{table*}

\paragraph{Recovery state machine}
The simulation skill compares torso height $h$ above the lowest body with the corresponding height $H$ of the default upright pose. It suspends the task prompt when $h<0.65H$ and restores it when $h>0.9H$. During recovery, every future slot requests the canonicalized default upright pose for all bodies, including torso orientation; velocity and style are masked.

The first slot's arrival time is
\begin{equation}
 \tau_0(h)=\operatorname{clip}\left(\frac{H-h}{v_{\mathrm{rise}}},
                      0.25\,\mathrm{s},1\,\mathrm{s}\right).
 \label{eq:recovery-arrival}
\end{equation}
Here, $v_{\mathrm{rise}}=0.8\,\mathrm{m/s}$ sets the requested arrival time, not the robot's actual vertical velocity. Each later slot adds $0.25\,$s, up to $2\,$s. On exit, the skill restores the suspended task prompt. Only the prompt changes between the task and recovery states.

\paragraph{G1 recovery}
The deployment skill follows the same height-triggered structure but requests only torso height, leaving torso planar position, all orientations, and limb targets unconstrained. Each of five future slots requests the calibrated upright torso height; velocity and style cues are masked. The first arrival time uses the height-gap rule in \cref{eq:recovery-arrival}, with an upper bound of $1.25\,$s rather than $1\,$s. Later slots add $0.25\,$s each, capped at $2\,$s.

Torso height is computed by FK relative to the lowest reconstructed body point. Recovery begins below $0.65H$ and ends above $0.90H$, where $H$ is the calibrated upright torso height. Dense tracking and unconditional generation disable automatic recovery. During headset-controlled crouching, height is measured relative to the lower foot and the entry threshold is based on the commanded torso height, bounded below by $0.65\,$m. The exit threshold remains $0.90H$, allowing intentional crouching while requiring recovery to finish upright.

\subsection{Deployment on the G1}
\label{sec:app-deploy}

The C++ deployment stack uses TensorRT with CUDA graphs on a laptop equipped with an NVIDIA RTX 5090 GPU. The measured latency from sensor acquisition to joint-command transmission is $1.3\,$ms; this is the sensor-to-command processing time, not the control period. An Xbox controller and a PICO headset supply runtime commands.

All hardware demonstrations use the deterministic prior mean: we set the sampling noise to zero, project $\mu_t$ onto the unit sphere, and decode it together with $d_t$. This differs from the sampled latent used during distillation. The deployment stack also supports stochastic sampling, but that option is not used for the reported hardware demonstrations.

G1 follows BeyondMimic~\cite{liao2026beyondmimic} for action representation. The setpoint is $q_j^{\mathrm{des}}=q_j^0+\alpha_j a_j$, where $\alpha_j=0.25\tau_j^{\max}/k_{p,j}$. These offsets are relative to the nominal pose, not the reference motion. The policy updates them every $20\,$ms (50 Hz), with four simulator physics steps per action. SOMA instead uses $q_j^{\mathrm{des}}=\pi\tanh(a_j)$ with fixed PD gains. Deployment must preserve joint order, offsets, scales, and gains. World-position skills additionally consume a localization estimate; canonicalized skills use the local pose convention above.

Control-mode selection updates the goal-related observations and masks while retaining the same policy weights and joint-control interface. For the sitting demonstrations, the robot is placed near the object and supplied with a nearby-object cue. This setup provides task context externally rather than estimating the object location or planning an approach autonomously.

\subsection{Evaluation Protocol}
\label{sec:app-reproduction}

Reference reconstruction and task execution evaluate different capabilities. Reconstruction uses the posterior with privileged reference information, whereas task and hardware evaluations use the prior. Task comparisons use matched initial states, commands, perturbations, horizons, and success criteria. The reference-tracking comparison excludes synthetic recovery data; the four-expert policy includes recovery training.

\fi

\end{document}